\documentclass[letterpaper, 10 pt, conference]{ieeeconf}  

\IEEEoverridecommandlockouts                              

\usepackage{graphicx}
\usepackage{tikz}
\usetikzlibrary{positioning, arrows.meta, calc,arrows}
\usetikzlibrary{patterns,decorations.pathmorphing}
\usetikzlibrary{patterns.meta, calc, angles}

\usepackage{amsfonts}

\usepackage{amsthm}
\usepackage{amsmath}
\usepackage{amssymb}
\usepackage{multirow}
\usepackage{algorithm}
\usepackage{algpseudocode}

\let\labelindent\relax
\usepackage{enumitem}
\usepackage{siunitx}

\usepackage{xcolor}
\usepackage{makecell}
\usepackage[inkscapelatex=false]{svg}
\usepackage{cite}

\definecolor{CaliforniaGold}{rgb}{0.992, 0.71, 0.082}
\definecolor{BerkeleyBlue}{rgb}{0.0, 0.149, 0.463}
\definecolor{darkblue}{rgb}{0.0, 0.13, 0.28}
\definecolor{fandango}{rgb}{0.71, 0.2, 0.54}
\definecolor{deepcarrotorange}{rgb}{0.91, 0.41, 0.17}
\definecolor{safepolytopegreen}{RGB}{51, 204, 51}

\usepackage{booktabs}
\usepackage{array}

\newtheorem{proposition}{Proposition}

\usepackage{hyperref}

\definecolor{codegreen}{rgb}{0,0.6,0}
\definecolor{codegray}{rgb}{0.5,0.5,0.5}
\definecolor{codepurple}{rgb}{0.58,0,0.82}
\definecolor{backcolour}{rgb}{0.95,0.95,0.95}
\definecolor{mycitecolor}{RGB}{71, 191, 38}
\definecolor{mylinkcolor}{RGB}{40, 115, 201}

\title{\LARGE \bf X-Safe: Embodiment-Agnostic Safeguarding for Robot Arm Manipulation Policies via Action Masking}

\author{Alex Beaudin$^*$, Hanna Krasowski$^*$, Kartik Nagpal$^*$, Sanjit A. Seshia, Murat Arcak, Negar Mehr
\thanks{$^*$ Equal contribution.}
\thanks{All authors are with the University of California, Berkeley, USA
        {\tt\small \{alex\_beaudin, krasowski, kartiknagpal, sseshia, arcak, negar\}@berkeley.edu}}%
}

\begin{document}

\maketitle
\thispagestyle{empty}
\pagestyle{empty}

\begin{abstract}

Ensuring safety of learning-enabled robotic manipulation across diverse embodiments and tasks still requires significant manual engineering.
For collision avoidance, existing approaches typically rely on embodiment-specific learned safety approximations or simplified robot models together with stopping-based fallback controllers that compromise performance.
In this paper, we propose an embodiment-agnostic safeguarding approach, \emph{X-Safe}, which provides formal probabilistic guarantees for collision avoidance without model simplifications or learned safety functions. By operating in the configuration space, our safeguarding method works with different types of robot policies. 
X-Safe reasons locally around the current configuration to establish safe regions and uses action masking to restrict the policy's support to these safe regions.
We demonstrate X-Safe on two embodiments and three policies, both in simulation and on hardware. We observe less than half of the performance degradation compared to state-of-the-art safeguarding, and empirically corroborate our formal collision avoidance guarantees.

\end{abstract}

\section{Introduction}\label{sec:introduction}
Safety in learning-based robotic systems is typically addressed through intrinsic alignment \cite{zhang2026safevla,brown21a-alignement} or test-time safeguarding \cite{brunke2022safe, krasowski2023provably}. While intrinsic alignment rewards safer policy behavior, test-time safeguarding provides the formal safety guarantees that are frequently required for real-world deployment. 
However, in the context of robotic manipulation, existing test-time safeguarding approaches typically need to be engineered for specific embodiments and tasks, e.g., by deriving simplified dynamics or collision models  \cite{Jung2025-RAIL, thumm2022provably}, or require gathering unsafe trajectories in simulation to learn representations of safety \cite{Yu2024-learnedCBF, tayal2026vocbf}. 
Additionally, fallback policies are usually stopping controllers, which tend to degrade performance~\cite{thumm2022provably, Shao2021}.
In practice, for high-dimensional action spaces and close-proximity tasks, e.g., in cluttered environments, existing safeguards struggle to scale to real-time and require tedious engineering~\cite{Jung2025-RAIL}.

In this paper, we propose a scalable and embodiment-agnostic safeguard that provides probabilistic guarantees for robotic manipulation. Our safeguard reasons about collision avoidance with respect to the robot's embodiment and quasi-static obstacles using an object-based scene representation, which describes the robot and obstacles spatially in the workspace. We achieve this by computing probabilistically safe polytopes in the robot configuration space, and use these polytopes for action masking to limit the support of the robot's policy. Further, we use the safe polytopes to define a fallback controller that moves the manipulator away from collisions. Importantly, our approach does not require data to learn safety representations or manual design of safety functions. Instead, our approach relies on collision-checking functions, which are readily available from simulation frameworks. 
Our safeguard is computationally efficient, which we demonstrate in simulation and on hardware for robotic manipulation tasks. Our main contributions are:
\begin{itemize}
    \item We propose \emph{X-Safe}, a framework in configuration space that safeguards manipulation policies and is transferable to different embodiments without safety function engineering or learning effort while ensuring probabilistic collision avoidance for quasi-static environments. 
    \item We achieve efficient computation by combining fast-computed safe polytopes with action masking.
    \item We compare our approach on robotic manipulators for pick-and-place tasks in cluttered quasi-static environments, empirically showing that collisions are below the expected probability.
\end{itemize}

\begin{figure*}
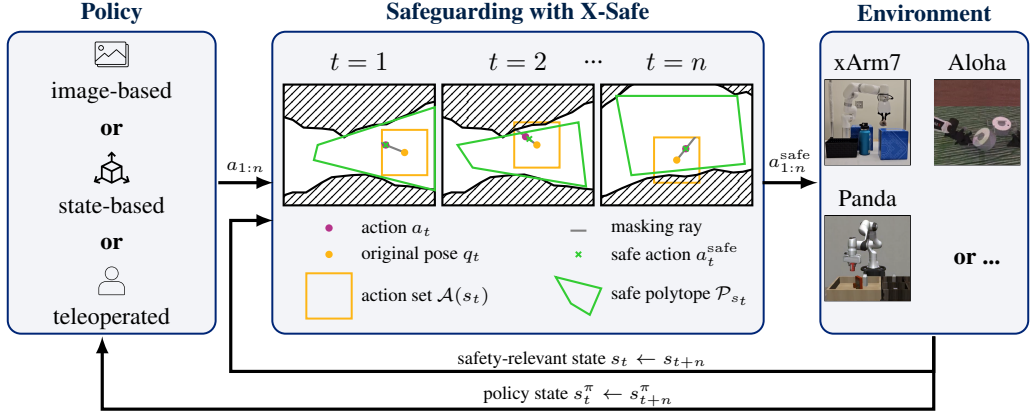

    \centering
    \vspace{0.2cm}
    \include{figs/headfigure.tex}
    \caption{Overview of our safeguarding method, X-Safe, that can take action sequences from policies together with a scene, i.e., safety-relevant state. The middle box illustrates safeguarding for $n$ actions in two exemplary directions of the configuration space. At $t=1$, the action set $\mathcal{A}$ is fully enclosed in the safe polytope $\mathcal{P}_s$ and consequently the action remains the same. In contrast, at $t=2$, the safe polytope intersects with the action set in the direction of the masking ray, and the proposed action $a_2$ is corrected to $a_2^\mathrm{safe}$. The safeguarding sequentially corrects all $n$ actions. }
    \label{fig:headfigure}
\end{figure*}

\section{Related Work}\label{sec:related_work}

\textbf{Safeguards with guarantees. }
Safeguarding methods for learning-based robotic systems have improved and expanded in the last decade \cite{brunke2022safe}. While the literature focuses on soft safety constraints, methods that can ensure probabilistic or hard guarantees on safety are necessary when a violation damages the robot or environment \cite{krasowski2023provably}. For continuous action spaces, methods that project unsafe actions to the boundary of a safe set dominate, e.g., using control barrier functions (CBFs) \cite{Li2019, Morton2025, wabersich2023data,sharma2024learning},
or reachability analysis \cite{Kochdumper2023, Selim2022a, Shao2021,bouvier2025ddat}. However, boundary projection frequently results in learning and deployment instabilities \cite{Markgraf2026, bejarano2024safety, Gros2020}. An alternative is continuous action masking, where the policy is extended with a function that ensures selecting safe actions only \cite{stolz2024excluding}. We introduce an action masking safeguard that corrects actions consistent with the policy's proposed direction in configuration space.

\textbf{Safeguarding for learned manipulation policies. }
An elementary but challenging notion of safety for robotic manipulation is collision avoidance with itself and the environment, e.g., static obstacles like tables or dynamic obstacles like humans \cite{Robey2026scienceopinion}. 
To achieve guarantees for learning-based policies, safeguards must additionally run in real time to effectively ensure safety, yet ideally minimally degrade performance. A common approach is to combine a monitor that detects unsafe actions with a fallback policy that replaces unsafe actions with stopping actions, compromising performance \cite{johansson2025safetyfilteringroboticmanipulation, thumm2025generalsafetyframeworkautonomous, thumm2022provably}. Another safeguarding solution is to learn CBFs from simulations or manually design candidate CBFs and use them in a quadratic program to filter unsafe actions \cite{Dastider2024-APEX, long2025neuralconfigurationspacebarriersmanipulation, Yu2024-learnedCBF}. However, candidate CBFs do not provide formal guarantees as they lack a proof of forward invariance \cite{kim2026safecontrolleractuallysafe}. Lastly, simplified robot models together with reachability analysis can be used to correct unsafe actions \cite{Jung2025-RAIL} but introduce conservatism that makes close-proximity operations challenging, e.g., grasping in cluttered environments. Instead, we propose a systematic, safe action set synthesis that is easily transferable across embodiments and environments as it relies only on an object-based scene representation and knowledge of the robotic arm's forward kinematics. 

\textbf{Safe trajectory optimization for manipulation. }
Even for traditional optimization or motion planning approaches, which usually require more information about the scene than learning-based approaches, ensuring action safety remains a complex challenge~\cite{smith2012dual}. The existence of many nonconvex obstacles in real-world robotic scenes often causes sampling-based planners to struggle finding feasible paths~\cite{orthey2023sampling, Elbanhawi2014} and optimization methods to converge to local minima and perform poorly~\cite{schulman2014motion, zucker2013chomp}. 
Mixed methods aim to decompose the free configuration space into large convex safe subsets to enable efficient, convex trajectory optimization \cite{deits2015computing, marcucci2023motion}. Recent research translated this concept to manipulation and proposed a real-time variant leveraging GPU parallelization \cite{werner2024faster, werner2025superfast}. Inspired by this line of work, we develop an efficient algorithm for local safe sets.

\section{Problem Formulation}\label{sec:problemformulation}
We denote sets and set operations with calligraphic letters. We let the function $\mathtt{vol}$ denote the Lebesgue measure for a set and $\langle A, b \rangle_\mathcal{P}$ abbreviate a polytope $\mathcal{P}$ in halfspace representation $\mathcal{P} = \{x \mid A x \leq b\}$. We define the safety-relevant state $s = (q, s^\mathrm{env}) \in \mathcal{S}$ comprised of a configuration $q \in \mathbb{R}^{DoF}$, with ${DoF}$ being the degrees of freedom of the manipulator, and the object-based scene representation $s^\mathrm{env}$. The scene representation $s^\mathrm{env}$ describes the pose and dimensions of objects in the environment apart from the robotic manipulator. The actions $a \in \mathcal{A}$ are defined relative to the current configuration of the robotic manipulator $q$, where the action set is $\mathcal{A}(s) \subset \mathbb{R}^{DoF}$. Intuitively, actions are viewed, locally, as relative changes in the configuration $q$ and clipped to $\mathcal{A}(s)$ based on a maximum joint difference $\Delta_{\max}$. Note that we omit the state-dependent notation for readability, and $\mathcal{A}^n$ is the action set for $n$ sequential actions. 

The violation function $C: \mathcal{S} \rightarrow \{0,1\}$ maps a state $s$ to $0$ if it is safe, and $1$ if a collision occurs. Thus, we can define the safe state-dependent set $\mathcal{C}_s^{\text{safe}} = \{a \mid C(s') = 0, s' = (q+a, s^\mathrm{env}), a \in \mathcal{A}\}$. 
Note that, in practice, the violation function $C$ is available through simulation frameworks, e.g., MuJoCo \cite{Todorov2012-Mujoco}. Specifically, $s^\mathrm{env}$ and $q$ are used to define a virtual twin of the physical environment, and forward kinematics allow for rapidly evaluating $C$ for configurations $q' = q+a$ that result from local actions $a$.
Let the policy $\pi$ take a state $s^\pi \in \mathcal{S}_\pi$ as argument and output action sequences in $\mathcal{A}^{n}$.
Note that the state space of the policy $\mathcal{S}_\pi$ can be different from the state space of the safeguard $\mathcal{S}$, e.g., a vision-based policy $\pi$ that uses images as state and vision-based perception that obtains $s^\mathrm{env}$ for the safeguard.

Given the violation function $C$, and a performant policy $\pi$ (i.e., near-optimal with high success rate), our goal is to construct a safe policy $\pi^\mathrm{safe}$ that satisfies a probabilistic safety guarantee. 
To that end, we define a safeguarding function $\mathcal{F}$ that wraps $\pi$ and satisfies
\begin{equation}\label{eq:desired_guarantee}
    \mathbb{P}\left[\mathcal{F}(\pi( s^\pi), s) \notin \mathcal{C}_s^\mathrm{safe}\right] \leq \delta, \text{ for all safe } s,
\end{equation}
where $\delta \in (0, 1]$ is a small tolerable probability of safety violation. 
In words, $\mathcal{F}$ modifies $\pi$ such that the resulting action is safe with high probability.
The probability in \eqref{eq:desired_guarantee} comes from $\mathcal{F}$, which may be stochastic, or simply work under limited information. 
Of course, $\mathcal{F} = 0$, i.e. stand still, satisfies \eqref{eq:desired_guarantee}, so another consideration for $\mathcal{F}$ is to leverage $\pi$.
The safe policy, then, is given by
\begin{equation}\label{eq:safe_policy}
\pi^{\mathrm{safe}}(s)  = \mathcal{F}(\pi(s^\pi), s).
\end{equation} 

\section{X-Safe: Probabilistic Safeguard for Manipulation}\label{sec:method}

Our safeguard provides a solution to \eqref{eq:desired_guarantee} by computing probabilistically safe polytopes locally in the configuration space (see Sec.~\ref{subsec:safepolytopes}) and defining an action correction function, called masking function, based on these polytopes (see Sec.~\ref{subsec:masking}). Fig.~\ref{fig:headfigure} summarizes our safeguard for action sequences. In the following, we discuss each of these components in more detail.

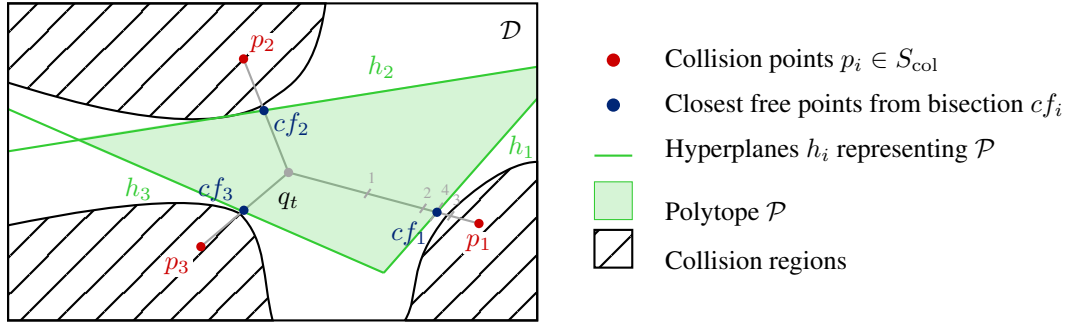
\begin{figure}[tb]
    \centering

\begin{tikzpicture}[scale=0.6]

    \colorlet{colorP}{red!80!black}          
    \colorlet{colorCF}{BerkeleyBlue}       
    \colorlet{colorH}{safepolytopegreen}         
    \colorlet{colorQLine}{gray!70}           
    \colorlet{colorAngleFill}{blue!20}       
    \colorlet{colorAngleDot}{blue!80!black}  
    \colorlet{colorQt}{black}                

    \def\W{10}
    \def\H{6}
    
    \coordinate (H1_A) at (0, 3.2);
    \coordinate (H1_B) at (\W, 5.0);
    \coordinate (H1_B2) at (\W, 4.8);
    
    \coordinate (H_int) at (7.3, 0.7); 
    \coordinate (H_int2) at (7.1, 0.9); 
    \coordinate (H3_A) at (0, 4.0);
    \coordinate (H2_B) at (\W, 4.2);
    \coordinate (H13_int) at (intersection of H1_A--H1_B and H3_A--H_int);

    \coordinate (CF1) at ($ (H1_A) ! 0.48 ! (H1_B) $);
    \coordinate (CF2) at ($ (H_int) ! 0.37 ! (H2_B) $);
    \coordinate (CF3) at ($ (H3_A) ! 0.60 ! (H_int) $);

    \coordinate (Q) at (5.3, 2.8);

    \coordinate (Safe1) at ($ (CF1) ! 0.1cm ! (Q) $);
    \coordinate (Safe2) at ($ (CF2) ! 0.2cm ! (Q) $);
    \coordinate (Safe3) at ($ (CF3) ! 0.1cm ! (Q) $);

    \coordinate (P1) at ($ (Q) ! 1.8 ! (CF3) $);
    \coordinate (P2) at ($ (Q) ! 1.2 ! (CF2) $);
    \coordinate (P3) at ($ (Q) ! 1.7 ! (CF1) $);

    \coordinate (mid_Q_CF2) at ($ (Q) ! 0.5 ! (CF2) $);
    \coordinate (mid_step2) at ($ (mid_Q_CF2) ! 0.5 ! (P2) $);
    \coordinate (mid_step3) at ($ (mid_step2) ! 0.5 ! (P2) $);
    \coordinate (mid_step4) at ($ (mid_step3) ! 0.5 ! (mid_step2) $);

    \draw[thick, pattern={Lines[angle=45, distance=10pt]}] 
        (0,\H) -- (6,\H) .. controls (5.8,4.5) and (5.2,4.4) .. 
        (CF1) .. controls (3.5,3.4) and (1.0,4.2) .. (0,4.5) -- cycle;

    \draw[thick, pattern={Lines[angle=45, distance=10pt]}] 
        (0,0) -- (5,0) .. controls (4.8,0.8) and (4.9,1.5) .. 
        (CF3) .. controls (3.5,2.5) and (1.0,2.0) .. (0,1.8) -- cycle;

    \draw[thick,pattern={Lines[angle=45, distance=10pt]}] 
        (7.5,0) -- (\W,0) -- (\W,3.0) .. controls (9.5,3.0) and (8.8,2.5) .. 
        (CF2) .. controls (7.5,1.0) and (7.8,0.5) .. (7.5,0);

    \draw[thick, colorQLine] (Q) -- (P1);
    \draw[thick, colorQLine] (Q) -- (P2);
    \draw[thick, colorQLine] (Q) -- (P3);

    

    \draw[thick, colorH] (H1_A) -- (H1_B2) node[pos=0.75, above left] {\normalsize $h_2$};
    \draw[thick, colorH] (H3_A) -- (H_int2) node[pos=0.45, below left, xshift=-0.2cm, yshift=0.2cm] {\normalsize $h_3$};
    \draw[thick, colorH] (H_int2) -- (H2_B) node[pos=0.8, below, xshift=0.2cm, yshift=0.1cm] {\normalsize $h_1$};

    \draw[draw=colorH, fill=colorH, opacity=0.2] (H13_int) -- (H1_B2) -- (H2_B) -- (H_int2) -- cycle;

    \draw[thick, colorQLine] ($ (mid_Q_CF2) ! 3pt ! 75:(CF2) $) -- ($ (mid_Q_CF2) ! 3pt ! -115:(CF2) $) node[anchor=south west, yshift=0.15cm, xshift=0.0cm, inner sep=1pt, text=colorQLine] {\tiny 1};
    \draw[thick, colorQLine] ($ (mid_step2) ! 3pt ! 75:(P2) $) -- ($ (mid_step2) ! 3pt ! -115:(P2) $)node[anchor=south west, yshift=0.15cm, xshift=0.0cm, inner sep=1pt, text=colorQLine] {\tiny 2};
    \draw[thick, colorQLine] ($ (mid_step3) ! 3pt ! 75:(P2) $) -- ($ (mid_step3) ! 3pt ! -115:(P2) $)node[anchor=south west, yshift=0.15cm, xshift=0.02cm, inner sep=1pt, text=colorQLine] {\tiny 3};
    \draw[thick, colorQLine] ($ (mid_step4) ! 6pt ! 75:(P2) $) -- ($ (mid_step4) ! 6pt ! -115:(P2) $)node[anchor=south west, yshift=0.3cm, xshift=0.1cm, inner sep=1pt, text=colorQLine] {\tiny 4};

    \draw[thick] (0,0) rectangle (\W,\H);
     \node at (9.5,5.5) {\normalsize $\mathcal{D}$};

    \fill[colorQLine] (Q) circle (2.5pt);
    \node[below, yshift=-4pt, text=colorQt] at (Q) {\normalsize $q_t$};

    \fill[colorP] (P1) circle (2.5pt);
    \node[below left=4pt, text=colorP,fill=white, inner sep=1pt] at (P1) {\normalsize $p_3$};

    \fill[colorP] (P2) circle (2.5pt);
    \node[below=3pt, text=colorP,fill=white, inner sep=1pt] at (P2) {\normalsize $p_1$};

    \fill[colorP] (P3) circle (2.5pt);
    \node[above right=2pt, text=colorP,fill=white, inner sep=1pt] at (P3) {\normalsize $p_2$};

    \fill[colorCF] (Safe1) circle (2.5pt);
    \node[below right=0pt, yshift=0.1cm, text=colorCF] at (Safe1) {\normalsize $cf_{2}$};

    \fill[colorCF] (Safe2) circle (2.5pt);
    \node[below left=0pt, text=colorCF] at (Safe2) {\normalsize $cf_{1}$};

    \fill[colorCF] (Safe3) circle (2.5pt);
    \node[above left=0pt, text=colorCF] at (Safe3) {\normalsize $cf_{3}$};

    \node[anchor=west, inner sep=5pt] at (-2.5, \H + 1.5) {\scriptsize
        \begin{tabular}{clcl}
            \tikz\fill[colorP] (0,0) circle (2.5pt); & Collision points $p_i \in S_\mathrm{col}$ 
           & \tikz\fill[colorCF] (0,0) circle (2.5pt); & Closest free points $cf_{i}$ \\[6pt]
            \tikz\draw[thick, pattern={Lines[angle=45, distance=10pt]}] (-0.25,-0.25) rectangle (0.25,0.25); & Collision regions & \tikz\draw[draw=colorH, fill=colorH, fill opacity=0.2] (-0.25,-0.25) rectangle (0.25,0.25); & Polytope $\mathcal{P}$ from hyperplanes $h_i$  \\
        \end{tabular}
    };

\end{tikzpicture}
    \caption{Illustration of $\mathtt{BisectToFree}$ and $\mathtt{PlaceNonRedundantHyperplanes}$. Sampled collision points $p_i$ for non-redundant hyperplanes in red, where for $p_1$ the bisection with a limit of 4 steps is illustrated with gray ticks. The points $cf_i$ are the last collision-free points. The hyperplanes $h_i$ are placed tangent to a sphere around $q_t$ with radius $cf_i$ and go through $cf_i$. The hyperplanes form the polytope $\mathcal{P}$.}
    \label{fig:bisect-hyperplane}
\end{figure}

\subsection{Safe Polytope Computation}\label{subsec:safepolytopes}

We assume that safety constraints encompass self-collision and collision with quasi-static environment objects. 
Specifically, the 
unsafe set $\mathcal{U}= \mathcal{U}^\mathrm{robot} \cup \mathcal{U}^\mathrm{env}$, where $\mathcal{U}^\mathrm{robot}$  describes  self-collisions, and $\mathcal{U}^\mathrm{env}$ describes environmental collisions, which can vary when the environment objects change, i.e., quasi-static setting.
Obtaining a closed-form representation of $\mathcal{U}$ is generally intractable in configuration space.
However, the violation function $C$, typically implemented with a simulation framework, can assess if a state $s \in \mathcal{U}$, in which case it returns $1$.

Thus, based on $C$, we aim to define an algorithm that provides configuration space sets, locally at the current configuration, that have only a low probability of intersection with $\mathcal{U}$. Specifically, we adapt the sampling-based method of Edge Inflation -- Zeroth Order (EI-ZO)  \cite{werner2024faster, werner2025superfast}. EI-ZO computes maximal probabilistically collision-free polytopes in the robot's configuration space to build a graph of convex sets, and we adapt it to an efficient algorithm for local safe polytopes, i.e., $\mathtt{LocalPolytope}$, in Alg.~\ref{alg:localZOSP}. 

\begin{algorithm}[tb]
    \caption{$\mathtt{LocalPolytope}$}\label{alg:localZOSP}
    \begin{algorithmic}[1]
    \Require Domain $\mathcal{D} \subseteq \mathbb{R}^{DoF}$, local configuration edge $(q_1, q_2)$ with $q_1, q_2 \in \mathcal{D}$, parameters $\varepsilon, \delta, \tau$, maximum number of iterations $N_\mathrm{iter}$, number of samples $M$.
    \Ensure Polytope $\mathcal{P} \subseteq \mathcal{D}$, Boolean $\texttt{passed}$ indicating if $\mathcal{P}$ satisfies \eqref{eq:desired_guarantee} for $(\varepsilon, \delta)$.
    \State $k \leftarrow 1$, $\mathcal{P} \leftarrow \mathcal{D}$
    \State $\mathtt{passed} \gets \texttt{False}$
    \For {$k = 1,\dotsc, N_\mathrm{iter}$}
        \State $S \sim \mathtt{UniformSample}(\mathcal{P}, M) \cup \mathcal{E}(q_1, q_2)$ 
        \State $S_{\text{col}} \leftarrow \mathtt{PointsInCollision}(S)$
        \State $\delta_k \gets \frac{6 \delta}{ \pi^2 k^2}$
        \If{$\mathtt{UnadaptiveTest}(\delta_k, \varepsilon, \tau, S_{\text{col}}, M)$ passes}
            \State $ \mathtt{passed} \gets \texttt{True}$
            \State \textbf{break}
        \EndIf
        \State $S^*_{\text{free}} \leftarrow \mathtt{BisectToFree}(S_{\text{col}}, q_1, q_2)$
        \State $\mathcal{P} \leftarrow \mathtt{PlaceNonRedundantHyperplanes}(\mathcal{P}, S^*_{\text{free}})$
        
    \EndFor
    \State \textbf{return} $\mathcal{P},\, \mathtt{passed}$
    \end{algorithmic}
\end{algorithm}

In essence, EI-ZO samples points in the configuration space, counts the number of points in collision $\mathtt{PointsInCollision}$, and checks if the number of colliding points is below a threshold defined by the statistical test $\mathtt{UnadaptiveTest}$. 
The test checks if there is enough data to have confidence $1 - \delta$ that the true fraction in collision is less than $\varepsilon$, given the observed samples.
If the test passes, then, in all likelihood, a probabilistically safe polytope has been found. 
Otherwise, the collision points refine the polytope by generating additional hyperplanes at boundary points between free space and collision found by bisection (see Fig.~\ref{fig:bisect-hyperplane}). Then new samples are drawn from $\mathcal{P}$, and the process iterates (more details see \cite{werner2025superfast}). 

To compute the probabilistic guarantees for the polytope generated by EI-ZO, we define an admissible volume fraction which allowed to be in collision $\varepsilon$ and a confidence threshold $\delta$. Further, we introduce a decision threshold $\tau$, which is typically set to $0.5$, in order to balance the power and cost of the statistical test. Consequently, the polytope returned by EI-ZO exhibits the probabilistic guarantee stated in Prop.~\ref{prop:action_single}.
\begin{proposition}[Safe polytope \cite{werner2024faster}]\label{prop:action_single}
    Let $\mathcal{C}_s^{\text{safe}}$ denote the true safe region at the current state $s$.
    Let an error tolerance $\varepsilon$, a confidence $\delta$, and a decision threshold $\tau$ be given. 
    Suppose that $\mathtt{LocalPolytope}$ returns $(\mathcal{P}_s, \mathtt{passed})$.
    Then, for any number of samples $M\geq \frac{2 \log(1/\delta)}{(\varepsilon\tau^2)}$, $\mathcal{P}_s$ satisfies
\begin{equation}
    \mathbb{P}\left[ \frac{\mathtt{vol}(\mathcal{P}_s \setminus \mathcal{C}_s^{\mathrm{safe}})}{\mathtt{vol}(\mathcal{P}_s)} > \varepsilon \right] \leq \delta.
\end{equation}
\end{proposition}
Intuitively, the polytope returned by the algorithm has a low probability, $\delta$, of containing many unsafe states (specifically, a fraction greater than $\varepsilon$).

Our main adaptations to EI-ZO are threefold. First, we add the extra samples $\mathcal{E}$ to the set of uniform configuration-space samples (Line 4, Alg.~\ref{alg:localZOSP}). In particular, the  $4 \times DoF$ extra samples $\mathcal{E}$ surround the local configuration edge $q_1$ and $q_2$ in axis-aligned directions (see Fig.~\ref{fig:extra_points}). Typically, the $q_1$ and $q_2$ of interest are the current configuration $q_t$ and candidate action configuration $q_t+a_t$.
Second, we speed up termination at the expense of slightly more conservative sets by placing hyperplanes at the last collision-free point along the direction defined by $q_1$ and $q_2$. As in EI-ZO, free points in a direction of interest are determined with bisection, but $\mathtt{BisectToFree}$ returns the last free instead of the last colliding point. 
Finally, the statistical test in EI-ZO (Line 7, Alg.~\ref{alg:localZOSP}) provides the probabilistic safety guarantee for the polytope stated in Prop.~\ref{prop:action_single} only if the algorithm terminates within the maximum number of iterations $N_\mathrm{iter}$. 
To detect if $N_\mathrm{iter}$ is sufficient to pass the statistical test, we return an additional Boolean value $ \mathtt{passed}$. 

\begin{figure}[tb]
    \centering
    \vspace{0.2cm}
\definecolor{docBlue}{RGB}{52, 152, 219}
\definecolor{docRed}{RGB}{231, 76, 60}

\begin{tikzpicture}[scale=0.7,>=stealth]

    \begin{scope}[shift={(0,1.8)}, thick]
        \draw[->] (0,0) -- (0.6,0) node[below] {$c_1$};
        \draw[->] (0,0) -- (0,0.6) node[left] {$c_2$};
    \end{scope}

    \coordinate (Q1) at (2.5, 2.5);
    
    \fill[docBlue] (Q1) circle (2.5pt);
    \node[docBlue, below right] at (Q1) {$q_1$};
    
    \fill[docRed] (1.2, 2.7) circle (2.5pt); 
    \fill[docRed] (2.5, 4.0) circle (2.5pt); 
    \fill[docRed] (2.5, 1.0) circle (2.5pt); 

    \coordinate (Q2) at (7.0, 2.5);
    \draw[dashed, docBlue, thick] (Q1) -- (Q2);

    \fill[docBlue] (Q2) circle (2.5pt);
    \node[docBlue, below left, xshift=-2pt] at (Q2) {$q_2$};
    
    \coordinate (R_top)    at (7.0, 4.0);
    \coordinate (R_bottom) at (7.0, 1.0);
    \coordinate (R_right)  at (8.6, 2.5);
    
    \fill[docRed] (R_top) circle (2.5pt);
    \fill[docRed] (R_bottom) circle (2.5pt);
    \fill[docRed] (R_right) circle (2.5pt);

    \draw[thick] (Q2) -- (7.0, 4.0);
    \draw[thick] (6.8, 4) -- (7.2, 4);
    
    \draw[thick] (Q2) -- (7.0, 1.0);
    \draw[thick] (6.8, 1.0) -- (7.2, 1.0);
    
    \draw[thick] (Q2) -- (8.6, 2.5);
    \draw[thick] (8.6, 2.3) -- (8.6, 2.7);
    \fill[docBlue] (7.0, 2.5)circle (2.5pt);
    \coordinate (R_top)    at (7.0, 4.0);
    \coordinate (R_bottom) at (7.0, 1.0);
    \coordinate (R_right)  at (8.6, 2.5);
    
    \fill[docRed] (R_top) circle (2.5pt);
    \fill[docRed] (R_bottom) circle (2.5pt);
    \fill[docRed] (R_right) circle (2.5pt);
    
    \node[above, font=\small] at (8, 2.6) {$\Delta_{\max}$};

    \node[anchor=center] at (4.8, 0.0) {$\mathcal{E} = \{ \textcolor{docRed}{\bullet} \mid \forall c_{\text{DoF}} \}$};

\end{tikzpicture}
    \vspace{-0.4cm}
    \caption{Set of extra samples $\mathcal{E}$ in configuration space surrounding the points of interest $q_1$ and $q_2$.}
    \label{fig:extra_points}
\end{figure}

\subsection{Action Masking Safeguard}\label{subsec:masking}
Given a safe polytope, our aim is to define a safeguarding function $\mathcal{F}^\mathrm{safe}$ such that only safe actions are executed.
In other words, they lead to the next configuration $q_{t+1}$ being within the safe polytope $\mathcal{P}_s$. 
To this end, we draw from ray masking in \cite{stolz2024excluding} to correct candidate actions $a$ consistent with the intended direction in configuration space to $a^\mathrm{safe}$. Effectively, ray masking radially contracts the support of the policy onto the safe polytope $\mathcal{P}_s$.
The intended direction connects the current configuration $q_t$ and the candidate action $a_t$, and is called "ray" (see Fig.~\ref{fig:headfigure}).
Note that since the current configuration $q_t$ is used as the origin of the rays, ray masking will only produce a stopping action if the candidate action $a_t$ is the stopping action. 

We summarize the adapted ray masking approach in Alg.~\ref{alg:raymask}. 
To efficiently determine if the ray direction requires action masking, we first check if the safe polytope $\mathcal{P}_s$ fully encloses the state-dependent action set $\mathcal{A}(s)$ (Line 1, Alg.~\ref{alg:raymask}) or the polytope encloses the action set in the ray direction (Line 5, Alg.~\ref{alg:raymask}). 
In both cases, the action will remain the same, i.e., $a_t = a_t^\mathrm{safe}$. 
If the action set intersects with the polytope in the ray direction, we proportionally move the actions towards the center on the ray (Line 9, Alg.~\ref{alg:raymask}). 
Intuitively, the closer the action is to the current configuration $q_t$, the smaller the correction (see Fig.~\ref{fig:headfigure}). 
Finally, note that projecting unsafe actions to the boundary of the safe polytope would be an alternative to masking \cite{krasowski2023provably, Markgraf2026}. 
However, for stochastic policies, projection places all unsafe actions on the polytope boundary \cite[Fig. 2]{krasowski2023provably}. 
Consequently, the manipulation policy would need to frequently navigate near-collision configurations, which is often challenging. In contrast, radially contracting the action distribution, as with ray masking, leads to boundary actions only if these actions were on the boundary of the action set before.

\begin{algorithm}[tb]
    \caption{X-Safe: Action Masking Safeguard}\label{alg:safeguard}
    \begin{algorithmic}[1]
    \Require Current configuration $q_t$, action set $\mathcal{A}$, domain $\mathcal{D}$, candidate actions $a_{1:n}$.
    \Ensure Safe actions $a^\mathrm{safe}_{1:n}$, each in $\mathcal{A}$.
    \State $q_1 \leftarrow q_t$
    \For {$i = 1,\dotsc, n$}
        \State $\mathcal{P}, \mathtt{passed} \gets \mathtt{LocalPolytope}(\mathcal{D}, (q_i, q_i+a_{i}), \varepsilon, \delta, N_\mathrm{iter}, M)$  \Comment{Alg.~\ref{alg:localZOSP}}
        \If {$\mathtt{passed}$}
            \State $ a^\mathrm{safe}_{i} \gets \mathtt{RayMask}(\mathcal{P}, \mathcal{A}, q_i, a_{i})$  \Comment{Alg.~\ref{alg:raymask}}
            \State $q_{i+1} \gets q_i + a^\mathrm{safe}_{i}$
        \Else 
            \State $a^\mathrm{safe}_{i:n} \gets \mathtt{ChebyshevFallback}(\mathcal{P}, q_i, \mathcal{A}, n-i+1)$ 
            \State \textbf{break}
        \EndIf
    \EndFor
    \State \textbf{return} $a^\mathrm{safe}_{1:n}$
    \end{algorithmic}
\end{algorithm}

Our safeguard based on action masking with safe polytopes is presented in Alg.~\ref{alg:safeguard}. 
Moreover, we can extend the probabilistic safety guarantee in Prop.~\ref{prop:action_single} for a single action to an action sequence as stated in Prop.~\ref{prop:action_sequence}. 
\begin{proposition}\label{prop:action_sequence}
    Let $a^\mathrm{safe}_{1:n}$ be a sequence of actions and $s_i$ the sequence of associated states computed from Alg.~\ref{alg:safeguard}.
    If $\mathtt{passed}$ is true for all $n$ actions, it holds that 
\begin{equation}
    \mathbb{P}\left[\frac{\mathtt{vol}(\mathcal{P}_{s_i} \setminus \mathcal{C}_{s_i}^{\mathrm{safe}})}{\mathtt{vol}(\mathcal{P}_{s_i})} > \varepsilon,  \text{\emph{for some }} s_i \right] \leq n \delta.
\end{equation}
\end{proposition}
The proof follows directly from the union bound. 
In words, the probability that any of the computed polytopes contains many collisions is low. 
Conversely, the probability that all of them have few collisions, i.e., less than $\varepsilon$, is high, meaning at least $1 - n\delta$.
For longer action sequences, our confidence in the safety of all polytopes degrades. 
Since policies are typically executed in a receding-horizon fashion with a short action sequence, this can be counteracted by a smaller $\delta$.

For a large number of iterations $N_\mathrm{iter}$, we expect $\mathtt{passed}$ to be true. 
Since more iterations may prohibitively increase runtime for real-time deployment, Alg.~\ref{alg:safeguard} introduces a fallback controller which moves the configuration to the Chebyshev center of the candidate polytope $\mathcal{P}$.
Specifically, the fallback controller $\mathtt{ChebyshevFallback}$ outputs an action chunk of actions with the same magnitude, where the last action realizes the Chebyshev center.
This fallback controller guides the manipulator away from unsafe configurations.

\subsection{Implementation}\label{subsec:implementation}

While our safeguard transfers easily to different embodiments and state-based as well as image-based policies, the core requirement is having access to the function $\mathtt{PointsInCollision}$, which evaluates the function $C$ for many configuration samples. In practice, this is widely accessible, e.g., when an object-based scene representation is obtained from a perception system, and the robotic manipulator is described by a Unified Robot Description Format (URDF) file~\cite{Todorov2012-Mujoco}. Further, we need to specify the domain $\mathcal{D}$, which is typically an axis-aligned box centered at the current configuration. For our experiments, we found that  $\mathcal{D} = \mathcal{A}$ based on maximum joint differences was most effective from both safety and performance standpoints.

For Alg.~\ref{alg:localZOSP}, a small $\delta$ leads to more iterations for convergence, and consequently increases runtime, but produces consistently better polytopes, even if $\mathtt{UnadaptiveTest}$ did not pass. 
Generally, setting a large number of maximum iterations $N_\mathrm{iter}$ is desired so that Alg.~\ref{alg:localZOSP} almost never reaches $N_\mathrm{iter}$ iterations and consequently Prop.~\ref{prop:action_sequence} empirically holds as well. 
Thus, a meaningful tuning approach is to first decrease $\delta$ until the polytope reaches the desired refinement level, and then decrease $N_\mathrm{iter}$ to improve the runtime.
For the parameter $\varepsilon$, we observed the least effects, and the chosen $0.02$ balances sampling costs and avoiding collisions well.

\section{Experiments}\label{sec:experiments}

We demonstrate X-Safe on two embodiments and four tasks for state-based and image-based policies in simulation and on hardware (see Fig.~\ref{fig:exp_overview}). 
We specify the safeguard parameters in Table~\ref{tab:parameters}.
We set $\mathcal{D} = \mathcal{A}$ and define the width of the box based on the maximum joint differences per single step $\Delta_\mathrm{max}$.
While seemingly innocuous, this domain choice has important effects on the ray masking.
If we set $\mathcal{D}$ larger than the maximum joint differences, then we will observe collisions farther in advance. 
If the set defined by maximum joint differences is fully enclosed in $\mathcal{P}_s$, masking will have no effect on the action.
Yet, more samples are required to adequately query the larger domain, at no additional benefit.
Specifically, we set $\Delta_\mathrm{max} = 0.015$ for the Franka Panda and the adversarial teleoperation on the UFactory xArm7, and $\Delta_\mathrm{max} = 0.03$ for UFactory xArm7 with a learned pick-and-place policy.

\begin{table}[tb]
\caption{Hyperparameters.}
\centering
\setlength{\tabcolsep}{4pt}
\begin{tabular}{ l c c}
    \toprule
    \textbf{Description} & \textbf{Name} & \textbf{Value} \\
    \midrule
    Num. Particles & $M$ & 1024 \\ 
    Max Iterations & $N_\mathrm{iter}$ & 12 \\ 
    Confidence & $\delta$ & 0.005 \\ 
    Fraction in Collision & $\varepsilon$ & 0.02 \\ 
    Power parameter & $\tau$ & 0.5 \\ 
    \bottomrule
\end{tabular}
  \label{tab:parameters}
\end{table}

\usetikzlibrary{calc, shapes.geometric, arrows.meta}

\newlength{\taskBigH}   \setlength{\taskBigH}{4.0cm}   
\newlength{\taskGap}    \setlength{\taskGap}{3pt}      
\newlength{\taskHeadH}  \setlength{\taskHeadH}{11pt}   
\newlength{\taskSmallW} \setlength{\taskSmallW}{2.9cm} 
\newlength{\taskLblW}   \setlength{\taskLblW}{1.5cm}   
\newlength{\taskSmallH} 
\newlength{\legPad}     \setlength{\legPad}{6pt}       
\newlength{\legW}       

\newcommand{\mkop}{0.8} 

\tikzset{
  taskimg/.style    = {inner sep=0pt, outer sep=0pt},
  overlaylbl/.style = {anchor=north west, xshift=3pt, yshift=-3pt,
                       fill=white, fill opacity=0.85, text opacity=1,
                       rounded corners=2pt, inner sep=2.5pt,
                       font=\footnotesize, align=left},
  sidelbl/.style    = {anchor=west, inner sep=0pt,
                       font=\footnotesize, align=left,
                       text width=\taskLblW},
  grouphead/.style  = {anchor=north west, inner sep=0pt, outer sep=0pt,
                       minimum height=\taskHeadH,
                       font=\footnotesize, align=left},
  mk/.style      = {draw=white, line width=0.6pt, inner sep=0pt, outer sep=0pt,
                    opacity=\mkop},
  startmk/.style = {mk, circle, fill=green!55!black, minimum size=7pt},
  goalmk/.style  = {mk, star, star points=5, star point ratio=2.3,
                    fill=red!85!black, minimum size=10pt},
  obstmk/.style  = {mk, regular polygon, regular polygon sides=3,
                    fill=orange!95!black, minimum size=12pt,
                    label={[text=black, inner sep=0pt, yshift=0pt,
                            opacity=\mkop, font=\tiny\bfseries]center:!}},
  robotmk/.style = {mk, diamond, fill=blue!70!black, minimum size=9pt},
  taskpath/.style= {white, densely dashed, line width=0.8pt,
                    -{Stealth[length=4pt]}, shorten <=4pt, shorten >=5pt},
  legendbox/.style = {draw=black!30, rounded corners=2pt, anchor=north west,
                      inner xsep=\legPad, inner ysep=3pt, outer sep=0pt,
                      font=\footnotesize},
}

\newcommand{\imgcoord}[4]{%
  \path ($(#2.south west)!#3!(#2.south east)$) coordinate (#1X)
        ($(#2.south west)!#4!(#2.north west)$) coordinate (#1Y);
  \coordinate (#1) at (#1X |- #1Y);%
}

\newcommand{\legsym}[1]{%
  \makebox[12pt]{\tikz[baseline={([yshift=-0.5ex]current bounding box.center)}]{#1}}}
\newsavebox{\legStartBox}
\newsavebox{\legGoalBox}
\newsavebox{\legObstBox}
\newsavebox{\legRobotBox}
\newsavebox{\legPathBox}
\newcommand{\legentry}[2]{\mbox{\usebox{#1}\hspace{3pt}#2}}

\begin{figure*}[!t]
    \centering
    \setlength{\taskSmallH}{\dimexpr(\taskBigH-\taskHeadH-2\taskGap)/2\relax}%
    \sbox{\legStartBox}{\legsym{\node[startmk]{};}}%
    \sbox{\legGoalBox}{\legsym{\node[goalmk]{};}}%
    \sbox{\legObstBox}{\legsym{\node[obstmk]{};}}%
    \sbox{\legRobotBox}{\legsym{\node[robotmk]{};}}%
    \sbox{\legPathBox}{\tikz[baseline=-0.5ex]\draw[taskpath, black!70,
        shorten <=0pt, shorten >=0pt] (0,0) -- (0.6,0);}%
    \begin{tikzpicture}
        \node[taskimg, anchor=north west] (a) at (0,0)
            {\includegraphics[height=\taskBigH]{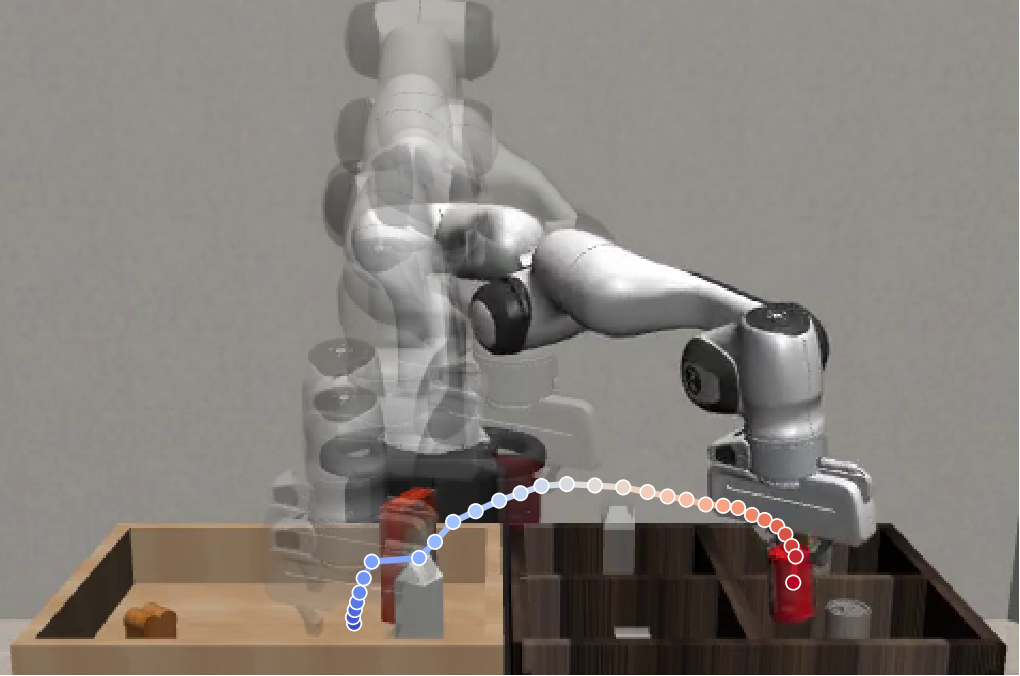}};
        \node[overlaylbl] at (a.north west)
            {\textbf{(a) Cluttered pick-and-place} \textit{(Panda)}};

        \imgcoord{aS}{a}{0.30}{0.15}   
        \imgcoord{aG}{a}{0.83}{0.15}   
        \imgcoord{aR}{a}{0.42}{0.4}    
        \draw[taskpath] (aS) to[bend left=35] (aG);
        \node[robotmk] at (aR) {};
        \node[startmk] at (aS) {};
        \node[goalmk]  at (aG) {};

        \node[taskimg, anchor=north west] (b) at ([xshift=\taskGap]a.north east)
            {\includegraphics[height=\taskBigH]{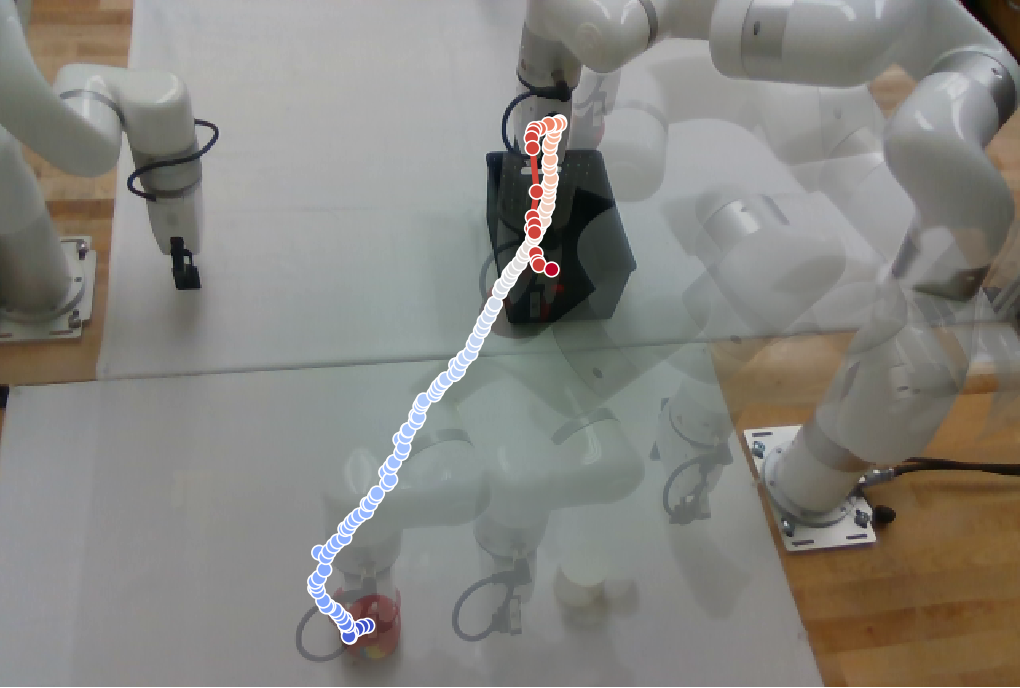}};
        \node[overlaylbl] at (b.north west)
            {\textbf{(b) Pick-and-place}\\\textbf{with obstacle}\\ \textit{(xArm7)}};

        \imgcoord{bS}{b}{0.35}{0.12}   
        \imgcoord{bG}{b}{0.52}{0.71}   
        \imgcoord{bO}{b}{0.56}{0.16}   
        \imgcoord{bR}{b}{0.66}{0.33}   
        \draw[taskpath] (bS) to[bend right=30] (bG);
        \node[robotmk] at (bR) {};
        \node[obstmk]  at (bO) {};
        \node[startmk] at (bS) {};
        \node[goalmk]  at (bG) {};

        \node[grouphead] (head) at ([xshift=2\taskGap]b.north east)
            {\textbf{(c) Adversarial Teleop.} \textit{(xArm7)}};

        \node[taskimg, anchor=north west] (c1)
            at ([yshift=-\taskGap]head.south west)
            {\includegraphics[width=\taskSmallW, height=\taskSmallH,
                              keepaspectratio]{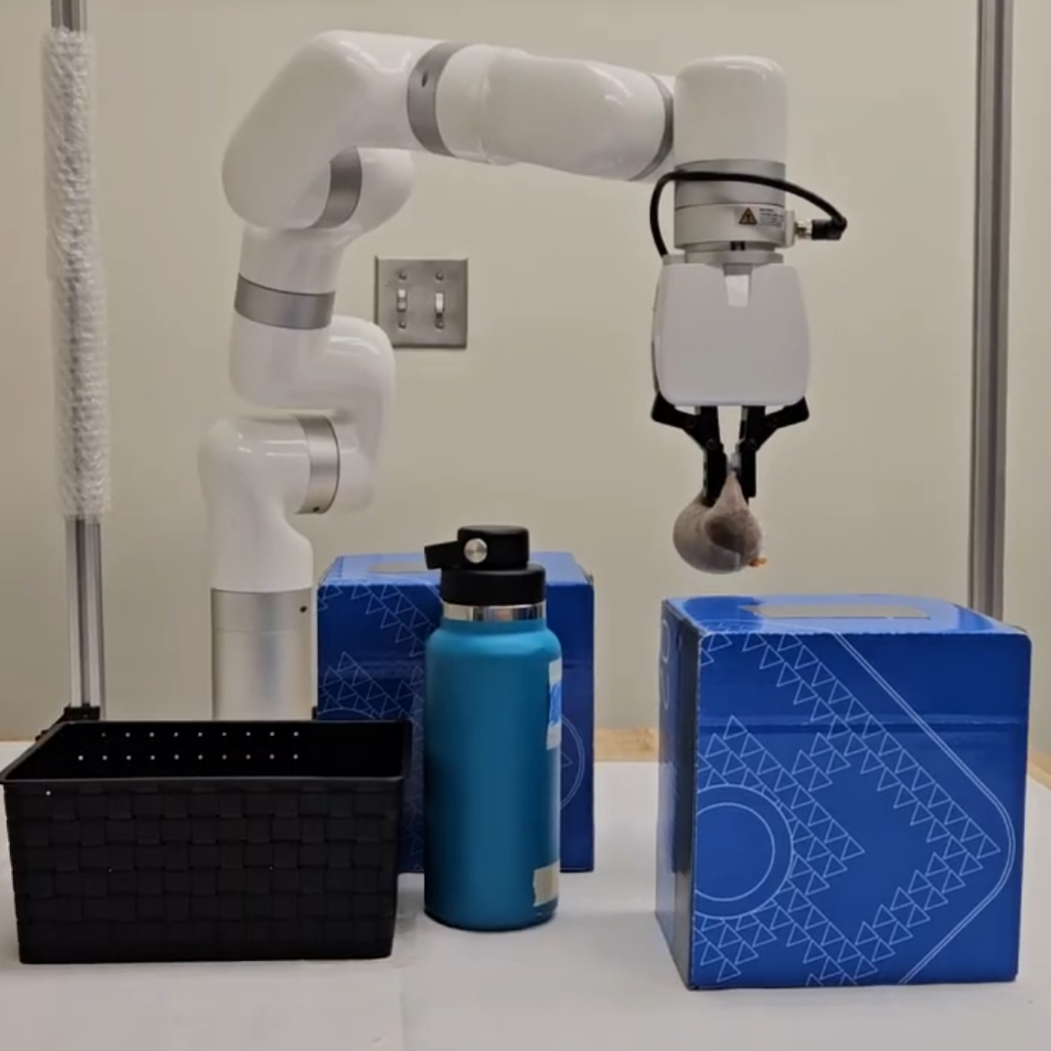}};
        \node[taskimg, anchor=south west] (c2)
            at ([xshift=2\taskGap]b.south east)
            {\includegraphics[width=\taskSmallW, height=\taskSmallH,
                              keepaspectratio]{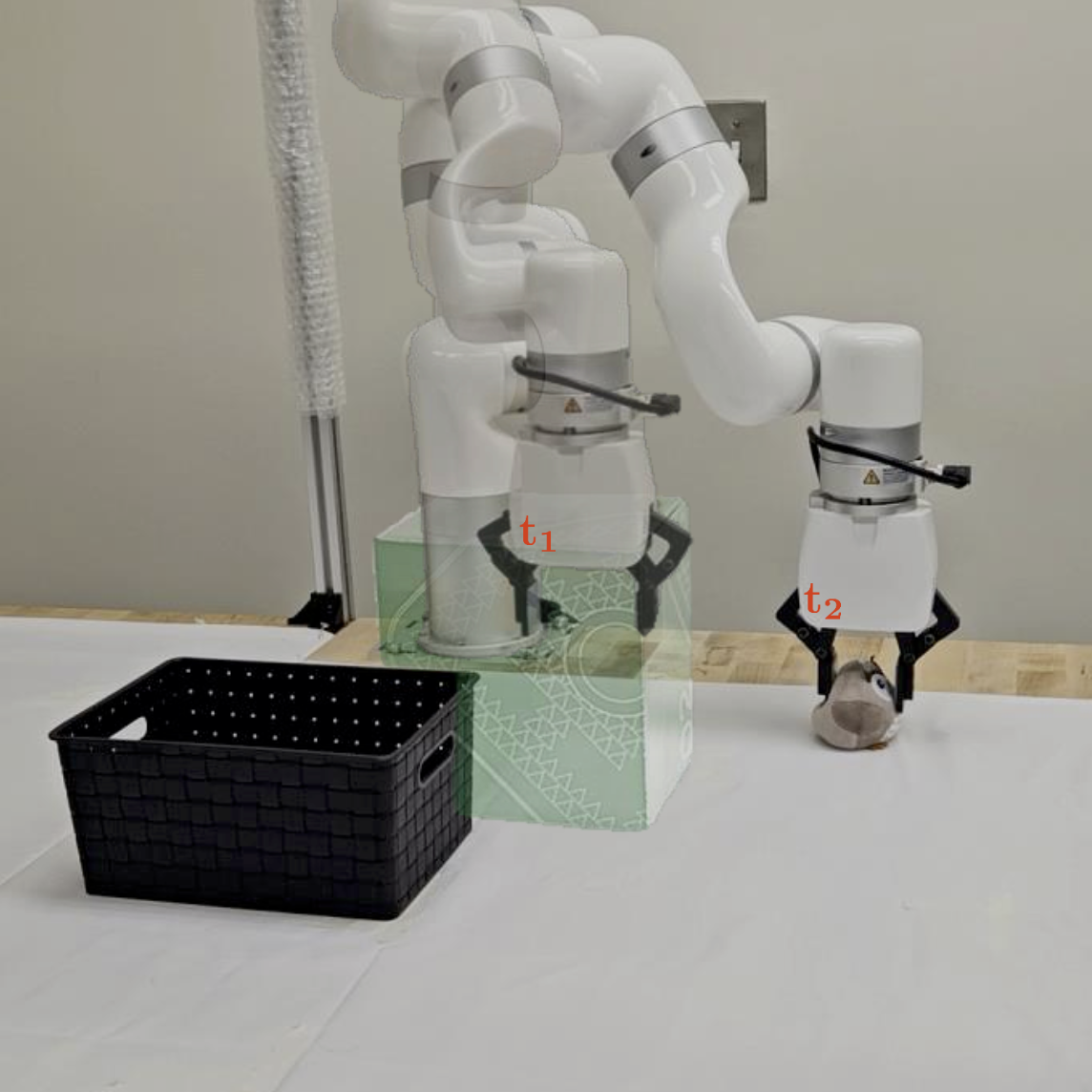}};

        \path let \p1=(c1.east), \p2=(c2.east) in
            ({max(\x1,\x2)+4pt},0) coordinate (lblx);
        \node[sidelbl] at (c1.west -| lblx)
            {\textbf{(c1)}\\Cluttered};
        \node[sidelbl] at (c2.west -| lblx)
            {\textbf{(c2)}\\Quasi-static};

        \path let \p1=(head.east), \p2=(lblx) in
            ({max(\x1,\x2+\taskLblW)},0) coordinate (cright);
        \draw[black!40, line width=0.4pt]
            (head.south west) -- (cright |- head.south west);

        \imgcoord{c1R}{c1}{0.7}{0.70}  
        \imgcoord{c1G}{c1}{0.22}{0.20}  
        \imgcoord{c1Oa}{c1}{0.5}{0.25} 
        \imgcoord{c1Ob}{c1}{0.82}{0.25} 
        \node[obstmk]  at (c1Oa) {};
        \node[obstmk]  at (c1Ob) {};
        \node[robotmk] at (c1R)  {};
        \node[goalmk]  at (c1G)  {};

        \imgcoord{c2S}{c2}{0.8}{0.32}  
        \imgcoord{c2R}{c2}{0.8}{0.52}  
        \imgcoord{c2G}{c2}{0.22}{0.32}  
        \imgcoord{c2O}{c2}{0.52}{0.45}  
        \node[obstmk]  at (c2O) {};
        \node[startmk] at (c2S) {};
        \node[goalmk]  at (c2G) {};
        \node[robotmk] at (c2R)  {};

        \pgfpointdiff{\pgfpointanchor{current bounding box}{south west}}
                     {\pgfpointanchor{current bounding box}{south east}}%
        \pgfgetlastxy{\figW}{\figDummy}%
        \setlength{\legW}{\dimexpr\figW-2\legPad\relax}%

        \node[legendbox] at ([yshift=-5pt]current bounding box.south west)
            {\makebox[\legW]{%
                \hfill\legentry{\legStartBox}{Object start position}%
                \hfill\legentry{\legGoalBox}{Object goal position}%
                \hfill\legentry{\legObstBox}{Obstacle}%
                \hfill\legentry{\legRobotBox}{Robot start}%
                \hfill\legentry{\legPathBox}{Nominal path}\hfill}};
    \end{tikzpicture}
    \caption{Our experimental tasks (and embodiments):
    (a)~Cluttered pick-and-place (Franka Panda).
    (b)~Pick-and-place with an obstacle (UFactory xArm7).
    (c)~Safeguarding during adversarial human teleoperation (xArm7) in
    (c1)~cluttered and (c2)~quasi-static environments.
    Together these span two embodiments, state- and image-based policies, and human operator control.}
    \label{fig:exp_overview}
\end{figure*}

\subsection{Franka Panda with state-based policy}
First, we compare X-Safe with RAIL \cite{Jung2025-RAIL}, a state-of-the-art safeguard that also provides formal collision guarantees for manipulation with learned policies without relying on a simple stopping fallback controller. Specifically, RAIL proposes a reachability approach, where an optimization problem is solved using a simplified model of the robotic manipulator to correct policy actions. 
Rather than providing empirical guarantees, e.g., as learned safety functions \cite{Yu2024-learnedCBF}, RAIL provides data-independent guarantees originating from the reachability analysis, resulting in a strong baseline from a safety standpoint.
We use the same task as RAIL, in which a Franka Panda manipulator needs to pick up a can among objects in one bin and drop it into another bin.
For each episode, the objects in the scene are randomly initialized (see  Fig.~\ref{fig:exp_overview}). For the policy, we use the state-based diffusion policy (DP) provided by \cite{Jung2025-RAIL}. All experiments are run on a AMD Ryzen 9 9950X3D 16-Core Processor and a Nvidia RTX 5090 32GB GPU. 

\begin{table}[tb]
\caption{Evaluation on different embodiments.} \vspace{-0.2cm}
\label{tab:results}
\centering
\setlength{\tabcolsep}{4pt}
\begin{tabular}{ l c c c }
\toprule
 \textbf{Method} & \textbf{Success (\%) $\uparrow$} & \textbf{Collision (\%) $\downarrow$}& \textbf{Horizon $\downarrow$}\\
\midrule
\multicolumn{4}{@{}l}{\textit{Franka Panda -- \includegraphics[width=0.4cm]{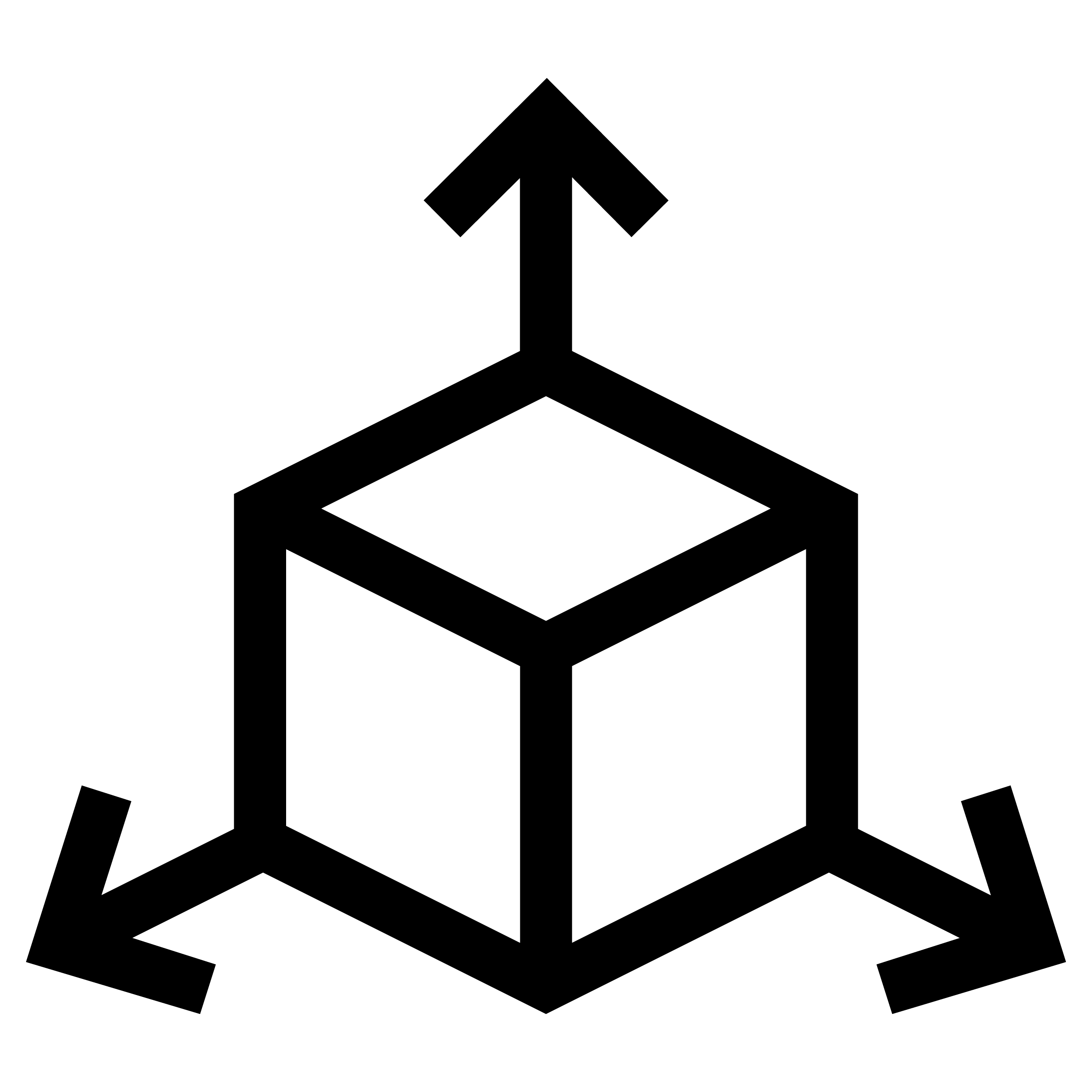} Pick and place a can in cluttered environment }}\\ \midrule
    DP    & $\mathbf{73.3 \pm 1.8}$ & $31.3 \pm 4.8$ & $\mathbf{911 \pm 28}$ \\
    RAIL  & $53.2 \pm 1.2 $ & $0.8 \pm 0.4$ & $1245 \pm 4$ \\
    DP + X-Safe & $66.4 \pm 1.3$ & $\mathbf{0.04 \pm 0.02}$  & $1057 \pm 17$ \\
\midrule
\multicolumn{4}{@{}l}{\textit{UFactory xArm7 --  \includegraphics[width=0.4cm]{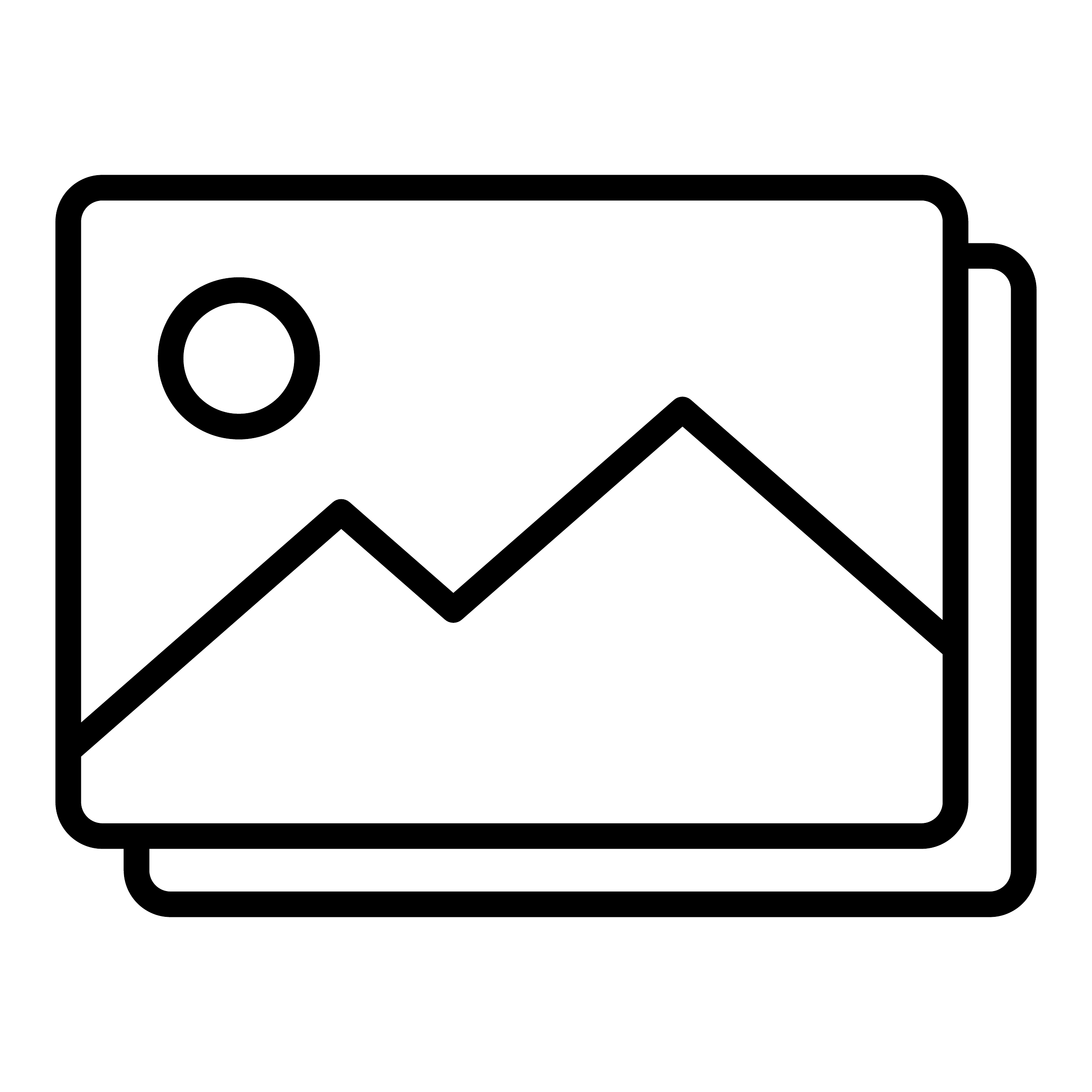} Pick and place into bin$^\dagger$}}\\ \midrule
    No obstacle & $\mathbf{64 \pm 9.6}$ & N/A$^*$ & $\mathbf{48.9 \pm 1.4}$ s \\
    Obstacle & $52 \pm 9.9$ & $100 \pm 0$ & $53.7 \pm 1.67$ s \\
    Obstacle + X-Safe & $44 \pm 9.9$ & $\mathbf{0 \pm 0}$ & $96.6 \pm 5.6$ s \\
\bottomrule
\end{tabular}
\vspace{0.1cm}
\\{\tiny Deviations indicate standard error in the mean. $^\dagger$ Horizon based on successful runs. \\ $^*$ No obstacle in the scene for the in-distribution baseline. 
}
\end{table}

Table~\ref{tab:results} presents our results. 
We report the task success rate per episode, the collision fraction of all time steps, and the average episode length. For the simulation experiment on the Franka Panda, we evaluated 200 episodes with maximum horizon of 2000 time steps and observed that the collision fraction with X-Safe is $0.02\%$---lower than the $2\%$ expected from the formal safety guarantee. While RAIL \cite{Jung2025-RAIL} reduces the performance of the DP by $20\%$ compared to the unsafe benchmark, the performance drop under X-Safe is only $7\%$. This is also reflected by the shorter average horizon for \mbox{X-Safe}. The intervention rate of X-Safe is $54.2 \pm 0.7$ percent of the time steps, which is higher than for RAIL. However, this is expected since action masking corrects actions when the free configuration space in the action direction is smaller than the action set, independent of the safety of the proposed action. Resulting in a look-ahead that essentially slows down the robot when approaching an unsafe region. In contrast, RAIL only intervenes when the proposed action sequence would lead to a collision, reflected by a lower intervention rate of $22.5 \pm 0.5$ percent. Lastly, we investigate the effect of the Chebyshev fallback controller, which is used on average for $4\%$ of the time steps. If we replace our fallback controller with a stopping fallback controller, the performance reduces from $66\%$ to $44\%$ as the arm deadlocks close to obstacles.

\subsection{UFactory xArm7 with image-based policy}
We evaluate two different settings on a UFactory xArm7. First, we safeguard an image-based DP for a pick-and-place task. In particular, we augment the original obstacle-free task with a cylindrical obstacle that intersects with the typical path of the DP policy (see  Fig.~\ref{fig:exp_overview}).
To train an imitation learning policy, we collected 100 rollout demonstrations via teleoperation. The architecture of the policy is a conditional diffusion transformer model \cite{dong2025mimic}. We use a standard score-matching reconstruction loss to train a visuomotor diffusion transformer that conditions on images and produces desired joint angles (in configuration space) and the gripper state.
To increase the safety margin on hardware, we check if the infinity norm of the action to the polytope boundary is less than $5 \times 10^{-3}$, and switch to the Chebyshev fallback controller in this case, i.e., even with polytopes that passed the statistical test.

For pick-and-place with xArm7, we evaluated 25 runs on hardware for each of the three settings (see Table~\ref{tab:results}). The baseline without the obstacles succeeds $64\%$ of the runs. Once the obstacle is introduced, the policy is operating out of the training distribution, which results in a performance drop to $52\%$ and collisions with the obstacle or table in every run. Specifically, the obstacle is hit in every run with varying severity (i.e., small contact forces to pushing the obstacle to fall over), and the table is hit in $8\%$ of the runs. When using X-Safe, the task is successfully executed in $44\%$ of the runs, and no collisions are observed. 
X-Safe trades 8 percentage points of success for the elimination of all collisions. The remaining task failures originate from a livelock: the fallback controller and the DP push in nearly opposite directions near the obstacle, loops away and toward the obstacle.
The increased time horizon for a run with X-Safe originates from slower operation around the obstacle. This is also reflected by the intervention rate of $71.2 \pm 1.4$ percent for successful runs, where interventions occur when the robot operates over the obstacle and to avoid table collisions before picking up the object.
In addition, we investigated the impact of perception uncertainty by calibrating camera-to-workspace estimates using homography. With the calibration, we estimated the obstacle’s position and inflated the obstacle by \SI{2}{\milli\meter}, which reflects the calculated calibration error. When re-running the experiment with X-Safe and the estimated position, we observed no collisions in 20 runs.

\subsection{UFactory xArm7 with adversarial teleoperation}
We use a 3DConnexion Spacemouse to guide the robot. The Spacemouse commands are given as absolute end-effector poses.
To extract joint configurations from these poses, we use the inverse kinematics solver from~\cite{sun2025dynamic}.
The target orientation is fixed to point down to simplify teleoperation.
These actions, which are the difference between the target configuration and current configuration, are then ray masked to yield $a^\mathrm{safe}$.
The resulting target configuration, $q_t + a^\mathrm{safe}_t$ is passed to the controller as a desired joint position.  

\begin{figure*}[tb]
    \centering
    \includegraphics[width=0.24\linewidth]{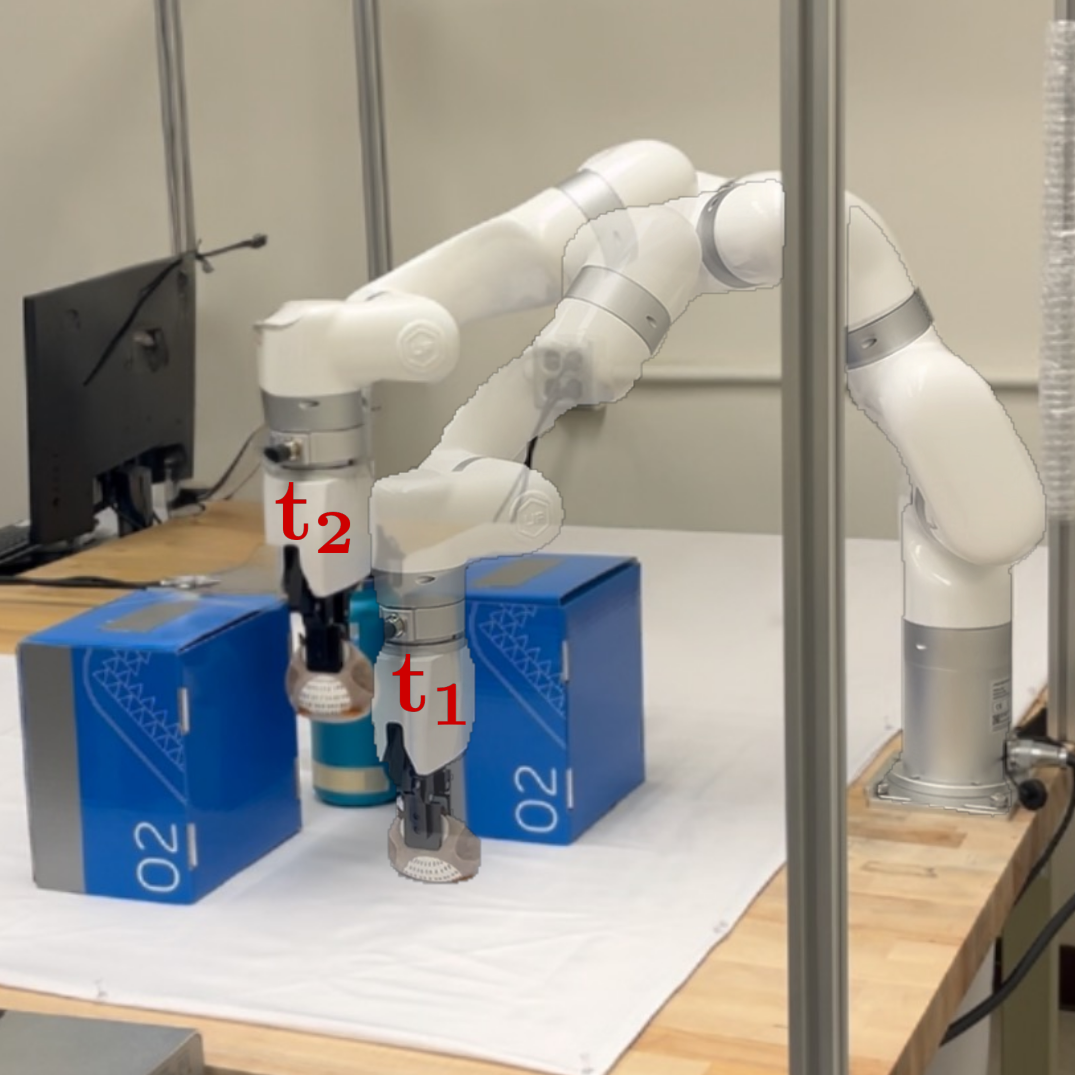}
    \includegraphics[width=0.24\linewidth]{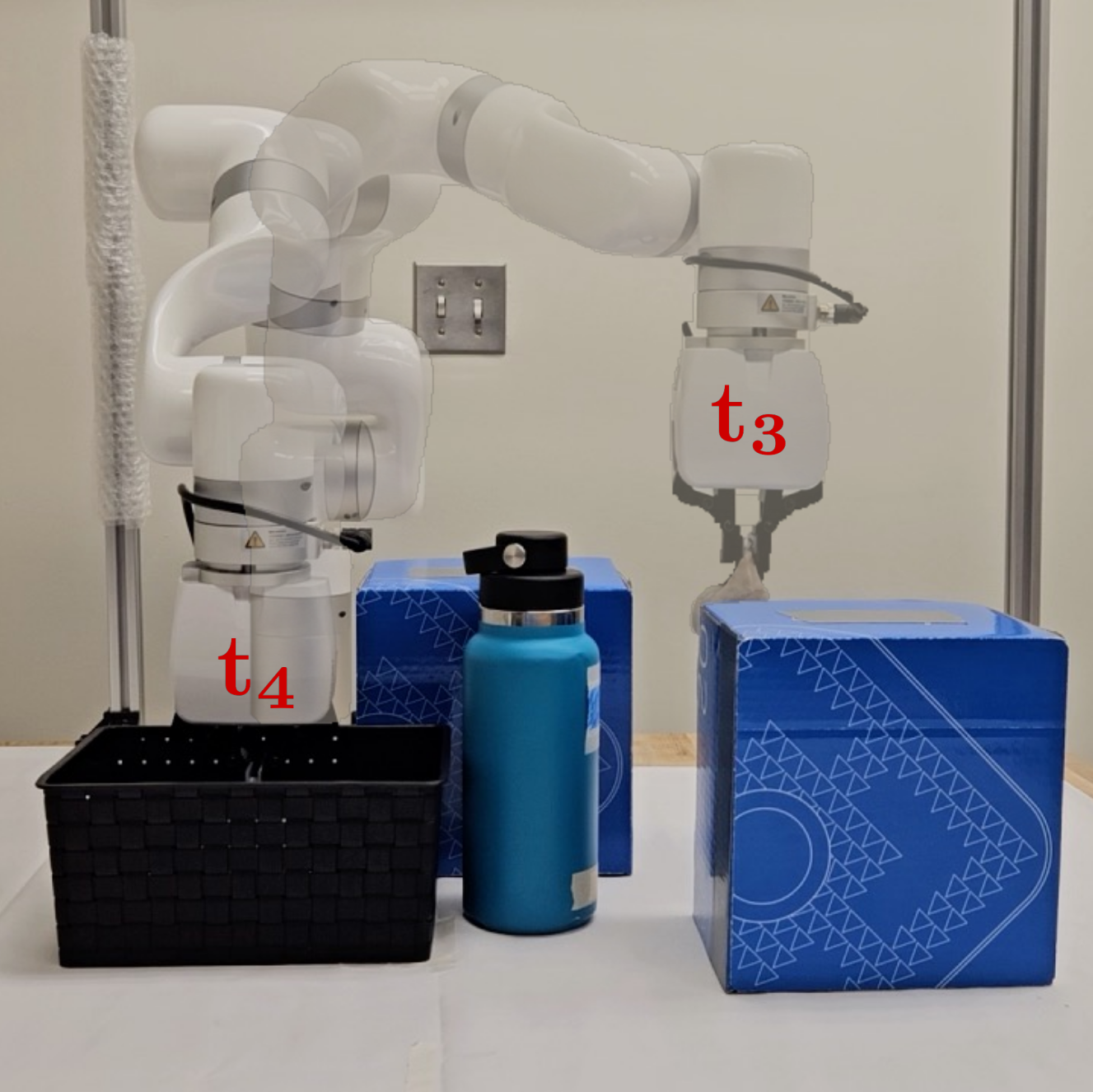} \hspace{0.1cm}
    \includegraphics[width=0.24\linewidth]{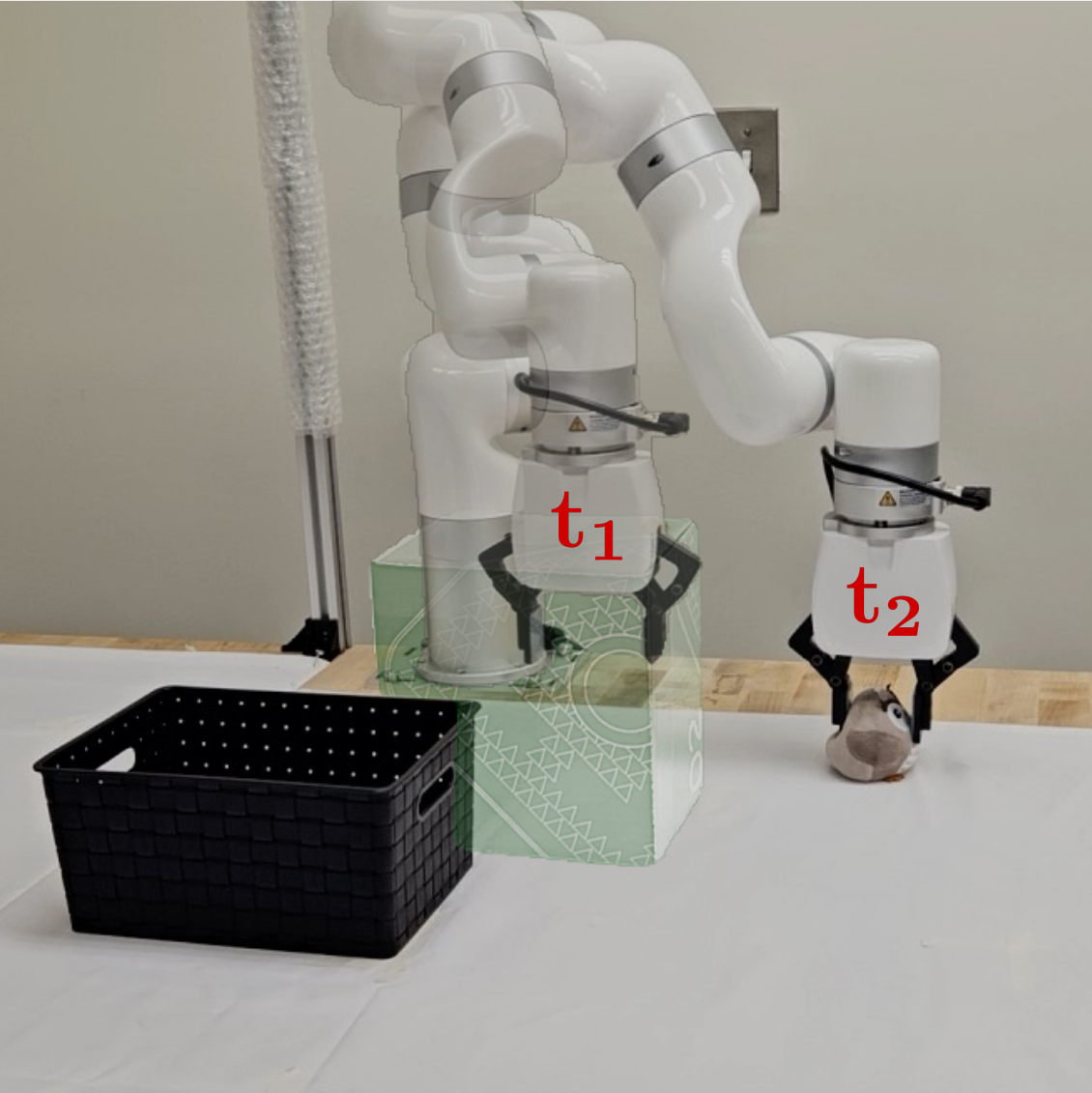}
    \includegraphics[width=0.24\linewidth]{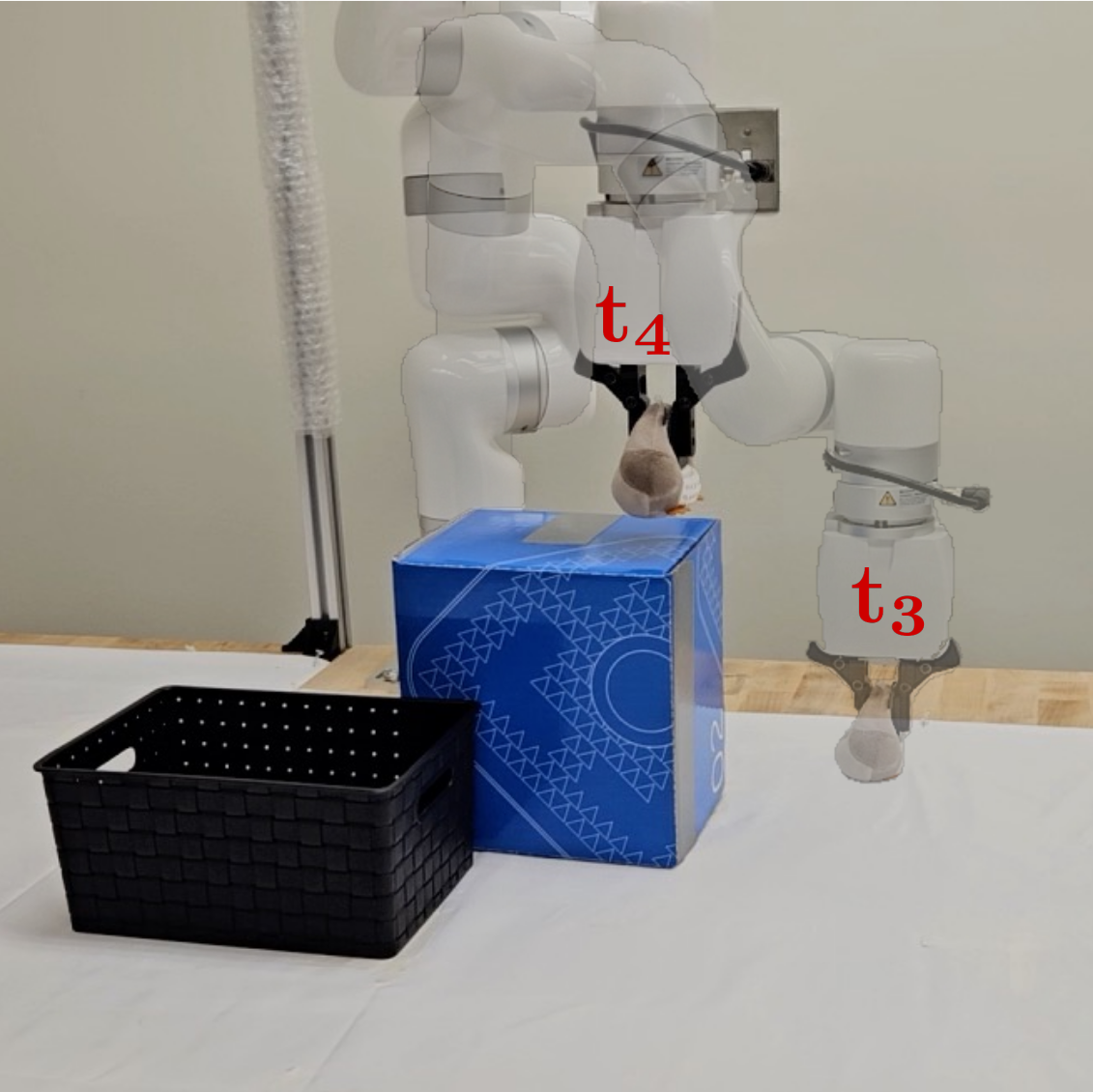} \\ 
    \includegraphics[clip, trim=0.5cm 0 0 0, width=0.49\linewidth]{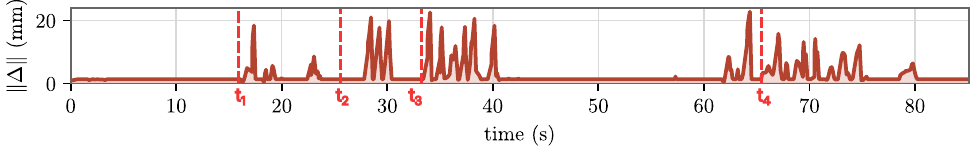} \hspace{0.1cm}
    \includegraphics[clip, trim=0.5cm 0 0 0, width=0.49\linewidth]{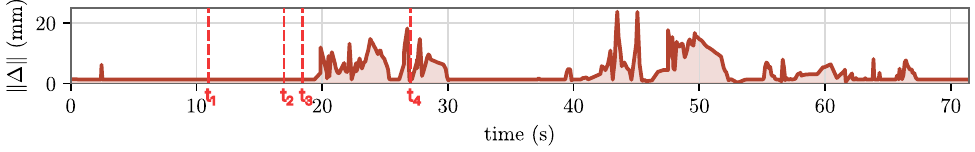}\\ \vspace{-0.2cm}
    {\footnotesize \textbf{(c1) Cluttered}} \hspace{7cm} {\footnotesize \textbf{(c2) Quasi-static}}
    \caption{Human-operated experiments. (c1) Cluttered. Human is pushing toward the left-most blue box at $t_1$ and arm stops at $t_2$; then at $t_3$ closely operating over a cluttered environment and dropping object into bin at $t_4$; (c2) Quasi-static. Human navigates to the object on direct path, i.e., through the obstacle that will appear at $t_3$ and is visualized in transparent green;  at $t_3$,  blue obstacle appears and blocks direct path.
    Bottom: The magnitude of the difference between target end effector positions before and after masking, in millimeters. }
    \label{fig:human_op_exp}
\end{figure*}

Fig.~\ref{fig:human_op_exp} illustrates the teleoperated manipulator experiments. The human operator adversarially moves the arm toward obstacles to demonstrate the tightness of X-Safe interventions. For the cluttered environment (i.e., Fig.~\ref{fig:human_op_exp}.c1), the manipulator can grasp between three objects.  When the adversarial human is trying to hit the obstacle, the Chebyshev fallback controller consistently moves the arm away from collision. Additionally, we present a quasi-static environment (i.e., Fig.~\ref{fig:human_op_exp}.c2). Once detected in the physical scene, an obstacle is added to the virtual environment simulated in parallel for the collision checking function $C$. Consequently, on the path to the bin, the adversarial operator can no longer take the shortest path because X-Safe prevents them from hitting the obstacle.

\section{Discussion}\label{sec:discussion}
The central assumption in X-Safe is the availability of the function $C$. To implement $C$, we use the simulation framework MuJoCo \cite{Todorov2012-Mujoco}, which allows us to easily represent different environments and embodiments, compute forward kinematics, and check for collisions. However, the integration becomes more involved for quasi-static environments, where perception has to detect safety-relevant environmental changes and translate them into simulation objects. Further, for dynamic obstacles, like humans, or more diverse notions of safety, like not spilling liquid, future work is needed to identify rapidly evaluable violation functions $C$. 

The safety guarantee provided in Prop.~\ref{prop:action_sequence} assumes uniform action selection from the safe polytope. Since only one action is selected from each polytope, the uniform assumption is reasonable, and our experimental evaluation confirmed this by producing lower collision rates than expected under the proposition. Additionally, the added samples $\mathcal{E}$ ensure a coverage surrounding the intended robot path, which supports detecting collisions in this relevant region. Nevertheless, further empirical validation would be beneficial to substantiate that the assumption is conservative on a broader set of tasks, policies, and embodiments.

We used the parameters reported in Table~\ref{tab:parameters} across tasks and embodiments to show that those can be set without significant tuning. The set $\mathcal{D}$ is varying according to $\Delta_{\mathrm{max}}$, where we took the $90$ percentile magnitude of actions, dimension-wise. 
The trade-off for $\mathcal{D}$ is that smaller reduces the number of samples required, while increasing the chance of adjusting more of the support of the policy. 
We defer the optimization $\mathcal{D}$ and other parameters to future work, as our goal is to demonstrate the ease of implementation and ability to transfer to other manipulators. 

Our experiments investigate the effect of X-Safe post-training for a limited number of tasks. We consistently observe a performance reduction of varying significance. This is expected as a posteriori safeguarding changes the distribution compared to the policy's training. To mitigate these distribution-shift effects, future work should fine-tune or train with X-Safe. Note that the differentiability of the function $\mathcal{F}^\mathrm{safe}$ allows for considering X-Safe as part of the environment, \cite{hunt2021verifiably, Markgraf2026} or as part of the policy \cite{stolz2024excluding}.

\section{Conclusion}\label{sec:conclusion}
We propose X-Safe, a safeguarding approach that provides probabilistic guarantees for collision avoidance of robot arm manipulators in quasi-static environments. Experiments on hardware and in simulation confirm the safety guarantees, which are provided even under adversarial operation in cluttered and quasi-static environments. At the same time, X-Safe shows significantly lower performance degradation relative to state-of-the-art safeguarding. Unlike existing approaches, X-Safe is transferable across embodiments with low tuning effort and without requiring additional training.






\bibliography{citations} 

@inproceedings{stolz2024excluding,
  title={Excluding the irrelevant: Focusing reinforcement learning through continuous action masking},
  author={Stolz, Roland and Krasowski, Hanna and Thumm, Jakob and Eichelbeck, Michael and Gassert, Philipp and Althoff, Matthias},
  booktitle={Advances in Neural Information Processing Systems},
  volume={37},
  pages={95067--95094},
  year={2024}
}

@article{bouvier2025ddat,
  title={Ddat: Diffusion policies enforcing dynamically admissible robot trajectories},
  author={Bouvier, Jean-Baptiste and Ryu, Kanghyun and Nagpal, Kartik and Liao, Qiayuan and Sreenath, Koushil and Mehr, Negar},
  journal={arXiv preprint arXiv:2502.15043},
  year={2025}
}

@inproceedings{sharma2024learning,
  title={Learning Differentiable and Safe Multi-Robot Control for Generalization to Novel Environments using Control Barrier Functions},
  author={Sharma, Vivek and Mehr, Negar and Hovakimyan, Naira},
  booktitle={2024 IEEE 63rd Conference on Decision and Control (CDC)},
  pages={8423--8428},
  year={2024},
  organization={IEEE}
}

@inproceedings{thumm2022provably, 
  title={Provably safe deep reinforcement learning for robotic manipulation in human environments},
  author={Thumm, Jakob and Althoff, Matthias},
  booktitle={2022 International Conference on Robotics and Automation (ICRA)},
  pages={6344--6350},
  year={2022},
  organization={IEEE}
}

@inproceedings{werner2025superfast,
    AUTHOR    = {Peter Werner AND Richard Cheng AND Tom Stewart AND Russ Tedrake AND Daniela Rus}, 
    TITLE     = {Superfast Configuration-Space Convex Set Computation on {GPUs} for Online Motion Planning}, 
    BOOKTITLE = {Proceedings of Robotics: Science and Systems}, 
    YEAR      = {2025}, 
    DOI       = {10.15607/RSS.2025.XXI.045} 
}

@article{werner2024faster,
  title={Faster Algorithms for Growing Collision-Free Convex Polytopes in Robot Configuration Space},
  author={Werner, Peter and Cohn, Thomas and Jiang, Rebecca H and Seyde, Tim and Simchowitz, Max and Tedrake, Russ and Rus, Daniela},
  journal={arXiv preprint arXiv:2410.12649},
  year={2024}
}

@article{krasowski2023provably,
  title={Provably safe reinforcement learning: Conceptual analysis, survey, and benchmarking},
  author={Krasowski, Hanna and Thumm, Jakob and M{\"u}ller, Marlon and Sch{\"a}fer, Lukas and Wang, Xiao and Althoff, Matthias},
  journal={Transactions on Machine Learning Research},
  year={2023}
}

@article{wabersich2023data,
  title={Data-driven safety filters: Hamilton-jacobi reachability, control barrier functions, and predictive methods for uncertain systems},
  author={Wabersich, Kim P and Taylor, Andrew J and Choi, Jason J and Sreenath, Koushil and Tomlin, Claire J and Ames, Aaron D and Zeilinger, Melanie N},
  journal={IEEE Control Systems Magazine},
  volume={43},
  number={5},
  pages={137--177},
  year={2023},
  publisher={IEEE}
}

@inproceedings{hunt2021verifiably,
  title={Verifiably safe exploration for end-to-end reinforcement learning},
  author={Hunt, Nathan and Fulton, Nathan and Magliacane, Sara and Hoang, Trong Nghia and Das, Subhro and Solar-Lezama, Armando},
  booktitle={Proceedings of the 24th International Conference on Hybrid Systems: Computation and Control},
  pages={1--11},
  year={2021}
}

@article{dong2025mimic,
  title={MIMIC-D: Multi-modal Imitation for MultI-agent Coordination with Decentralized Diffusion Policies},
  author={Dong, Dayi and Bhatt, Maulik and Choi, Seoyeon and Mehr, Negar},
  journal={arXiv preprint arXiv:2509.14159},
  year={2025}
}

@article{Robey2026scienceopinion,
author = {Alexander Robey  and Zachary Ravichandran  and Eliot Krzysztof Jones  and Jared Perlo  and Fazl Barez  and Vijay Kumar  and J. Zico Kolter  and Hamed Hassani  and George J. Pappas },
title = {Beyond alignment: {Why} robotic foundation models need context-aware safety},
journal = {Science Robotics},
volume = {11},
number = {113},
year = {2026},
doi = {10.1126/scirobotics.aef2191},
}

@inproceedings{Selim2022a,
  author = {Selim, Mahmoud and Alanwar, Amr and El-Kharashi, M. Watheq and Abbas, Hazem M. and Johansson, Karl H.},
  booktitle = {Proc. of the Int. Conf. on Communications, Signal Processing, and their Applications (ICCSPA)},
  pages = {1--6},
  title = {Safe Reinforcement Learning using Data-Driven Predictive Control},
  year = {2022},
}

@article{Kochdumper2023,
  author = {Kochdumper, Niklas and Krasowski, Hanna and Wang, Xiao and Bak, Stanley and Althoff, Matthias},
  journal = {IEEE Open Journal of Control Systems},
  pages = {79--92},
  title = {Provably Safe Reinforcement Learning via Action Projection Using Reachability Analysis and Polynomial Zonotopes},
  volume = {2},
  year = {2023},
}

@article{Gros2020,
  author = {Gros, Sebastien and Zanon, Mario and Bemporad, Alberto},
  journal = {IFAC-PapersOnLine},
  number = {2},
  pages = {8076--8081},
  title = {Safe Reinforcement Learning via Projection on a Safe Set: How to Achieve Optimality?},
  volume = {53},
  year = {2020},
}

@article{Li2019,
  author = {Li, Xiao and Serlin, Zachary and Yang, Guang and Belta, Calin},
  journal = {Science Robotics},
  number = {37},
  title = {A formal methods approach to interpretable reinforcement learning for robotic planning},
  volume = {4},
  year = {2019},
}

@article{bejarano2024safety,
  title={Safety filtering while training: {Improving} the performance and sample efficiency of reinforcement learning agents},
  author={Bejarano, Federico Pizarro and Brunke, Lukas and Schoellig, Angela P},
  pages = {788--795},
  volume = {10},
  number = {1},
  journal={IEEE Robotics and Automation Letters},
  year={2025},
}

@article{brunke2022safe,
  title={Safe learning in robotics: From learning-based control to safe reinforcement learning},
  author={Brunke, Lukas and Greeff, Melissa and Hall, Adam W and Yuan, Zhaocong and Zhou, Siqi and Panerati, Jacopo and Schoellig, Angela P},
  journal={Annual Review of Control, Robotics, and Autonomous Systems},
  volume={5},
  pages={411--444},
  year={2022},
  publisher={Annual Reviews}
}

@article{Markgraf2026,
  title = {Safe Reinforcement Learning using Action Projection: {Safeguard} the Policy or the Environment?},
  author = {Markgraf, Hannah and Sawant, Shambhuraj and Krasowski, Hanna and Schäfer, Lukas and Gros, Sebastien and Althoff, Matthias},
  year = {2026},
  journal = {Transactions on Machine Learning Research},
}

@INPROCEEDINGS{Jung2025-RAIL,
  author={Jung, Wonsuhk and Anthony, Dennis and Mishra, Utkarsh A. and Arachchige, Nadun Ranawaka and Bronars, Matthew and Xu, Danfei and Kousik, Shreyas},
  booktitle={IEEE International Conference on Robotics and Automation (ICRA)}, 
  title={{RAIL: Reachability}-Aided Imitation Learning for Safe Policy Execution}, 
  year={2025},
  pages={3582--3589},
  doi={10.1109/ICRA55743.2025.11128656}}

@INPROCEEDINGS{Yu2024-learnedCBF,
  author={Yu, Mingxin and Yu, Chenning and Naddaf-Sh, M-Mahdi and Upadhyay, Devesh and Gao, Sicun and Fan, Chuchu},
  booktitle={IEEE International Conference on Robotics and Automation (ICRA)}, 
  title={Efficient Motion Planning for Manipulators with Control Barrier Function-Induced Neural Controller}, 
  year={2024},
  volume={},
  number={},
  pages={14348--14355},
  doi={10.1109/ICRA57147.2024.10610785}}

@article{johansson2025safetyfilteringroboticmanipulation,
      title={Safety filtering of robotic manipulation under environment uncertainty: a computational approach}, 
      author={Anna Johansson and Daniel Lindmark and Viktor Wiberg and Martin Servin},
      year={2025},
      journal={arXiv preprint 2509.12674},
}

@INPROCEEDINGS{Dastider2024-APEX,
  author={Dastider, Apan and Fang, Hao and Lin, Mingjie},
  booktitle={IEEE/RSJ International Conference on Intelligent Robots and Systems (IROS)}, 
  title={{APEX: Ambidextrous} Dual-Arm Robotic Manipulation Using Collision-Free Generative Diffusion Models}, 
  year={2024},
  pages={9526--9533},
  doi={10.1109/IROS58592.2024.10802655}}

@article{thumm2025generalsafetyframeworkautonomous,
      title={A General Safety Framework for Autonomous Manipulation in Human Environments}, 
      author={Jakob Thumm and Julian Balletshofer and Leonardo Maglanoc and Luis Muschal and Matthias Althoff},
      year={2025},
      journal={arXiv preprint 2412.10180},
}

@article{long2025neuralconfigurationspacebarriersmanipulation,
      title={Neural Configuration-Space Barriers for Manipulation Planning and Control}, 
      author={Kehan Long and Ki Myung Brian Lee and Nikola Raicevic and Niyas Attasseri and Melvin Leok and Nikolay Atanasov},
      year={2025},
      journal={arXiv preprint 2503.04929},
}

@article{tayal2026vocbf,
    title={V-{OCBF}: Learning Safety Filters from Offline Data via Value-Guided Offline Control Barrier Functions},
    author={Mumuksh Tayal and Manan Tayal and Aditya Singh and Shishir Kolathaya and Ravi Prakash},
    journal={Transactions on Machine Learning Research},
    issn={2835-8856},
    year={2026},
    url={https://openreview.net/forum?id=PGO9mpIyyb},
    note={}
}

@inproceedings{zhang2026safevla,
title={Safe{VLA}: Towards Safety Alignment of Vision-Language-Action Model via Constrained Learning},
author={Borong Zhang and Yuhao Zhang and Jiaming Ji and Yingshan Lei and Josef Dai and Yuanpei Chen and Yaodong Yang},
booktitle={The Thirty-ninth Annual Conference on Neural Information Processing Systems},
year={2026},
url={https://openreview.net/forum?id=dt940loCBT}
}

@InProceedings{brown21a-alignement,
  title = 	 {Value Alignment Verification},
  author =       {Brown, Daniel S and Schneider, Jordan and Dragan, Anca and Niekum, Scott},
  booktitle = 	 {Proceedings of the 38th International Conference on Machine Learning},
  pages = 	 {1105--1115},
  year = 	 {2021},
}

@ARTICLE{Shao2021,
  author={Shao, Yifei Simon and Chen, Chao and Kousik, Shreyas and Vasudevan, Ram},
  journal={IEEE Robotics and Automation Letters}, 
  title={{Reachability-Based Trajectory Safeguard (RTS): A} Safe and Fast Reinforcement Learning Safety Layer for Continuous Control}, 
  year={2021},
  volume={6},
  number={2},
  pages={3663--3670},
  doi={10.1109/LRA.2021.3063989}}

@INPROCEEDINGS{Morton2025,
  author={Morton, Daniel and Pavone, Marco},
  booktitle={IEEE/RSJ International Conference on Intelligent Robots and Systems (IROS)}, 
  title={Safe, Task-Consistent Manipulation with Operational Space Control Barrier Functions}, 
  year={2025},
  pages={187--194},
  doi={10.1109/IROS60139.2025.11246389}}

@inproceedings{deits2015computing,
  title={Computing large convex regions of obstacle-free space through semidefinite programming},
  author={Deits, Robin and Tedrake, Russ},
  booktitle={Algorithmic Foundations of Robotics XI: Selected Contributions of the Eleventh International Workshop on the Algorithmic Foundations of Robotics},
  pages={109--124},
  year={2015},
}

@article{marcucci2023motion,
  title={Motion planning around obstacles with convex optimization},
  author={Marcucci, Tobia and Petersen, Mark and Von Wrangel, David and Tedrake, Russ},
  journal={Science robotics},
  volume={8},
  number={84},
  year={2023},
}

@article{orthey2023sampling,
  title={Sampling-based motion planning: {A} comparative review},
  author={Orthey, Andreas and Chamzas, Constantinos and Kavraki, Lydia E},
  journal={Annual Review of Control, Robotics, and Autonomous Systems},
  volume={7},
  year={2023},
}

@ARTICLE{Elbanhawi2014,
  author={Elbanhawi, Mohamed and Simic, Milan},
  journal={IEEE Access}, 
  title={Sampling-Based Robot Motion Planning: A Review}, 
  year={2014},
  volume={2},
  pages={56--77},
  doi={10.1109/ACCESS.2014.2302442}}

@article{smith2012dual,
  title={Dual arm manipulation — {A} survey},
  author={Smith, Christian and Karayiannidis, Yiannis and Nalpantidis, Lazaros and Gratal, Xavi and Qi, Peng and Dimarogonas, Dimos V and Kragic, Danica},
  journal={Robotics and Autonomous systems},
  volume={60},
  number={10},
  pages={1340--1353},
  year={2012}
}

@article{schulman2014motion,
  title={Motion planning with sequential convex optimization and convex collision checking},
  author={Schulman, John and Duan, Yan and Ho, Jonathan and Lee, Alex and Awwal, Ibrahim and Bradlow, Henry and Pan, Jia and Patil, Sachin and Goldberg, Ken and Abbeel, Pieter},
  journal={The International Journal of Robotics Research},
  volume={33},
  number={9},
  pages={1251--1270},
  year={2014},
}

@article{zucker2013chomp,
  title={Chomp: {Covariant} hamiltonian optimization for motion planning},
  author={Zucker, Matt and Ratliff, Nathan and Dragan, Anca D and Pivtoraiko, Mihail and Klingensmith, Matthew and Dellin, Christopher M and Bagnell, J Andrew and Srinivasa, Siddhartha S},
  journal={The International journal of robotics research},
  volume={32},
  number={9-10},
  pages={1164--1193},
  year={2013},
}

@misc{kim2026safecontrolleractuallysafe,
      title={Is Your Safe Controller Actually Safe? A Critical Review of CBF Tautologies and Hidden Assumptions}, 
      author={Taekyung Kim},
      year={2026},
      eprint={2603.06954},
      archivePrefix={arXiv},
      primaryClass={cs.RO},
      url={https://arxiv.org/abs/2603.06954}, 
}

@article{sun2025dynamic,
  title={Dynamic Rank Adjustment in Diffusion Policies for Efficient and Flexible Training},
  author={Sun, Xiatao and Yang, Shuo and Chen, Yinxing and Fan, Francis and Liang, Yiyan and Rakita, Daniel},
  journal={arXiv preprint arXiv:2502.03822},
  year={2025}
}

@INPROCEEDINGS{Todorov2012-Mujoco,
  author={Todorov, Emanuel and Erez, Tom and Tassa, Yuval},
  booktitle={2012 IEEE/RSJ International Conference on Intelligent Robots and Systems}, 
  title={MuJoCo: A physics engine for model-based control}, 
  year={2012},
  volume={},
  number={},
  pages={5026-5033},
  doi={10.1109/IROS.2012.6386109}}
\bibliographystyle{IEEEtran}

\appendix

\begin{algorithm}[H]
    \caption{$\mathtt{RayMask}$}\label{alg:raymask}
    \begin{algorithmic}[1]
    \Require Safe polytope $ \mathcal{P}_s =\langle A, b \rangle_{\mathcal{P}}$, state-dependent action set $\mathcal{A}$ with vertices $v_\mathcal{A}$, current configuration $q$, candidate action $a \in \mathcal{A}$ .
    \Ensure Safe action $a^\mathrm{safe} \in \mathcal{P}_s$
    \If{$A \, v_\mathcal{A} - b \leq 0$}
    \State $a^\mathrm{safe} = a$  \Comment{Safe polytope $\mathcal{P}_s$ encloses action set $\mathcal{A}$}
    \Else
        \State $q^\mathcal{A}_a = \mathtt{compute\_A\_boundary}(a, \mathcal{A})$  \Comment{Identifying for closest boundary of $\mathcal{A}$ to ray}
        \If{$A \, q^\mathcal{A}_a - b \leq 0$}
        \State $a^\mathrm{safe} = a$ \Comment{Action set $\mathcal{A}$ encloses $\mathcal{P}_s$ in direction of candidate action}
        \Else
        \State $q^\mathcal{P}_a = \mathtt{compute\_P\_boundary}(a, \mathcal{P}_s)$  \Comment{Identifying for closest halfspace of $\mathcal{P}_s$ to ray}
        \State $a^\mathrm{safe} = \frac{\|q^{\mathcal{P}}_a - q\|}{\|q^\mathcal{A}_a - q\|} \, a$  \Comment{Adapted ray masking \cite{stolz2024excluding}}
        \EndIf
    \EndIf
    \State \textbf{return} $a^\mathrm{safe}$
    \end{algorithmic}
\end{algorithm}

\end{document}